\documentclass[preprint,twocolumn,5p]{elsarticle}
\usepackage{amssymb}
\usepackage{lineno}

\usepackage{amsmath,amsfonts}
\usepackage{algorithm}
\usepackage{subfigure}
\usepackage[dvipsnames]{xcolor}
\usepackage{amssymb}
\usepackage{amsmath}
\usepackage[hidelinks]{hyperref}

\usepackage{amsmath}
\usepackage{amsthm}

\usepackage{algpseudocode}

\newtheorem{definition}{Definition}
\newtheorem{lemma}{Lemma}
\newtheorem{proposition}{Proposition}

\newcommand{\mypara}[1]{\paragraph{{\normalfont\textbf{#1}}}}
\newcommand{\yxzhu}[1]{\textcolor{black}{#1}}
\newcommand{\chen}[1]{\textcolor{black}{#1}}

\journal{Neural Networks}

\begin{document}

\begin{frontmatter}

\title{On the Probability of Necessity and Sufficiency of Explaining Graph Neural Networks: A Lower Bound Optimization Approach}

\tnotetext[1]{This research was supported in part by National Key R\&D Program of China (2021ZD0111501), National Science Fund for Excellent Young Scholars (62122022), Natural Science Foundation of China (61876043, 61976052, 62206064), the major key project of PCL (PCL2021A12).}

\author[Address1,Address2]{Ruichu Cai}
\ead{cairuichu@gmail.com}

\author[Address1]{Yuxuan Zhu\corref{cor1}}
\ead{iamyuxuanzhu@gmail.com}
\cortext[cor1]{Corresponding author}

\author[Address1]{Xuexin Chen}
\ead{im.chenxuexin@gmail.com}

\author[Address3]{Yuan Fang}
\ead{yfang@smu.edu.sg}

\author[Address4]{Min Wu}
\ead{wumin@i2r.a-star.edu.sg}

\author[Address1,Address2]{Jie Qiao}
\ead{qiaojie.chn@gmail.com}

\author[Address5]{Zhifeng Hao}
\ead{haozhifeng@stu.edu.cn}

\address[Address1]{School of Computer Science, Guangdong University of Technology, Guangzhou 510006, China}
\address[Address2]{Peng Cheng Laboratory, Shenzhen 518066, China}
\address[Address3]{School of Computing and Information Systems, Singapore Management University, 178902, Singapore}
\address[Address4]{Institute for Infocomm Research (I$^2$R), A*STAR, 138632, Singapore}
\address[Address5]{College of Science, Shantou University, Shantou 515063, China}

\begin{abstract}
The explainability of Graph Neural Networks (GNNs) is critical to various GNN applications, yet it remains a significant challenge. A convincing explanation should be both necessary and sufficient simultaneously. However, existing GNN explaining approaches focus on only one of the two aspects, necessity or sufficiency, or a heuristic trade-off between the two. Theoretically, the Probability of Necessity and Sufficiency (PNS) holds the potential to identify the most necessary and sufficient explanation since it can mathematically quantify the necessity and sufficiency of an explanation. Nevertheless, the difficulty of obtaining PNS due to non-monotonicity and the challenge of counterfactual estimation limit its wide use. To address the non-identifiability of PNS, we resort to a lower bound of PNS that can be optimized via counterfactual estimation, and propose a framework of Necessary and Sufficient Explanation for GNN (NSEG) via optimizing that lower bound. Specifically, we depict the GNN as a structural causal model (SCM), and estimate the probability of counterfactual via the intervention under the SCM. Additionally, we leverage continuous masks with a sampling strategy to optimize the lower bound to enhance the scalability. Empirical results demonstrate that NSEG outperforms state-of-the-art methods, consistently generating the most necessary and sufficient explanations.
The implementation of our NSEG is available at \url{https://github.com/EthanChu7/NSEG}.
\end{abstract}


% \begin{highlights}
% \end{highlights}


\end{frontmatter}

% \linenumbers

\section{Introduction}
Graph Neural Networks (GNNs) differentiate themselves from neural networks designed for Euclidean data by not only learning feature information but also capturing graph structures through the message-passing mechanism \cite{c:gcn,c:garphsage,c:gin,c:gat,c:link_pred,c:predict_n_propogate,c:mp_chem,c:sim_gnn,c:spectural_graph}. This unique characteristic has facilitated the successful application of GNNs in various domains, including social recommendation \cite{c:gnnforsocialrecommendation,c:knowgnnsocialrecomend}, molecule discovery \cite{c:graphaf,a:gcn_drug}, and fraud detection \cite{c:loan_fraud_dect,c:upfn_gnn}.
However, GNNs with high complexity are still considered black-box models \cite{a:explainable_ai/ArrietaRSBTBGGM20,b:interpretableML}, which limits their applications in many real-life related domains like medicine and healthcare \cite{a:bio_example}. 
Although numerous studies have been proposed to explain neural networks for Euclidean data \cite{c:lime,c:shap,c:xai_aximatic_att,c:causal_shap}, such approaches are usually not suitable for GNNs as they cannot explain the graph structures well. 
Hence, the explainability of GNNs remains an open challenge.

\begin{figure}
    \centering
    \includegraphics[scale=0.42]{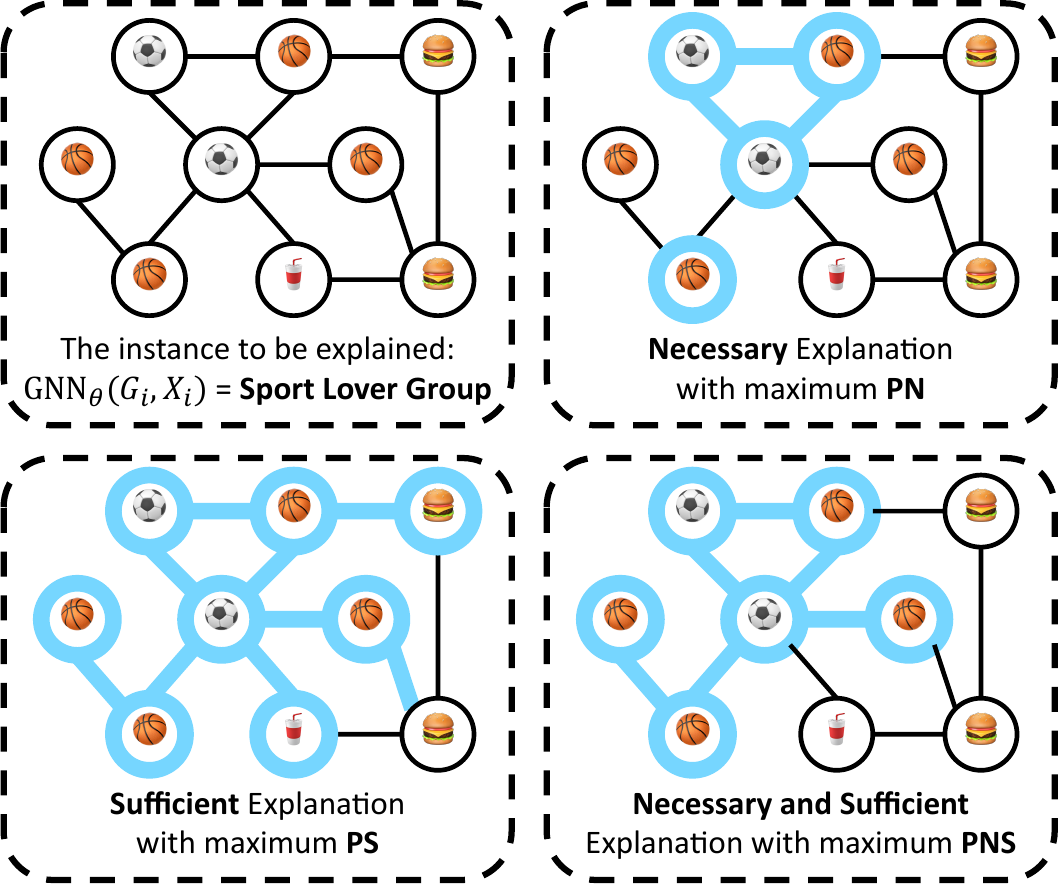}
    \caption{Explanations for the prediction ``Sport Lover Group", which are highlighted in \textcolor{SkyBlue}{blue}. Each node is a member in the group whose node features are their hobby, denoted by the icons. PN, PS, PNS refer to the probability of necessity, the probability of sufficiency, and the probability of necessity and sufficiency, respectively.}
    \label{fig:intro}
\end{figure}

Current approaches for explaining GNNs mainly search for three types of explanation, i.e., necessary explanation, sufficient explanation, and the heuristic trade-off explanation between necessity and sufficiency.
First, the approaches \cite{c:gem,a:rcexplainer,c:cf_gnnexplainer} searching for a necessary explanation seek to identify a group of necessary features that will change the prediction if one performs a perturbation. Although necessity is important for the explanation, the lack of sufficiency can result in the incompleteness of the explanations. For instance, consider a chat group classification task in which the hobby and social connections of each member are provided and we aim to explain why the instance given in Figure~\ref{fig:intro} is predicted as ``Sport Lover Group''. For the necessary explanation shown on the top-right of Figure~\ref{fig:intro}, only a small set of soccer lovers or basketball lovers are considered as explanations. However, it is insufficient as some basketball lovers are missing. In contrast, the approaches \cite{c:gnnexplainer,c:pgexplainer} searching for a sufficient explanation seek to locate a subset of the graph that can sufficiently cause the outcome by maximizing the mutual information between the input and outcome. For such approaches, the sufficient explanation might not be concise enough for people to understand, e.g., in the bottom-left of Figure~\ref{fig:intro}, the explanation covers almost the whole graph including both sports and snack lovers and their relationships. Furthermore, a recent approach \cite{c:cf2} considers a trade-off between necessity and sufficiency, whereas the trade-off is heuristically determined by hyper-parameters so that the explanation obtained might not be the \emph{most} necessary and sufficient.

Different from either sufficient or necessary explanations, a both necessary and sufficient explanation offers completeness without sacrificing conciseness. As illustrated in the bottom-right of Figure~\ref{fig:intro}, the most necessary and sufficient explanation includes all sport lovers, ensuring completeness, while excluding the redundant snack lovers, ensuring conciseness. The necessity and sufficiency of the explanation deserve a privileged position in the theory and practice of explainable AI \cite{c:pns_xai,a:pns}, and we argue that \emph{an ideal explanation should be most necessary and sufficient}.
A formal way to quantify the necessity and sufficiency of an explanation is through the use of the \textit{Probability of Necessity and Sufficiency} (PNS) \cite{b:causality}. However, there are two main challenges in identifying PNS: 1) the violation of the assumption of monotonicity making PNS not identifiable \cite{a:pns,c:pns_tian} and 2) the difficulty of counterfactual estimation. On one hand, to address the above identifiability issue of PNS, we derive a lower bound of PNS that can be estimated via counterfactual estimation. On the other hand, for counterfactual estimation, we depict the GNN as a structural causal model (SCM) and estimate the counterfactual probability by intervention under the SCM. In addition, to enable tractable optimization, continuous masks with a sampling strategy are used to optimize the lower bound of PNS. By combining these techniques, we propose a framework of \textbf{N}ecessary and \textbf{S}ufficient \textbf{E}xplanation for \textbf{G}NNs (NSEG), via maximizing the lower bound of PNS.

Our contributions can be summarized as follows:

\begin{itemize}
    \item We propose NSEG to generate necessary and sufficient explanations for GNNs, by optimizing a lower bound of PNS via counterfactual estimation.
    
    \item We depict the GNN as an SCM such that the counterfactual estimation is tractable via intervention on the SCM.
 We further leverage a continuous mask with a sampling strategy to optimize the lower bound of PNS, making the optimization tractable by relaxing the discrete explanation to a continuous case.
    
    \item Our experiments show that the explanations from NSEG are the most necessary and sufficient, and both aspects are critical to the generation of explanations.
    
\end{itemize}

\section{Related Work}

\subsection{Graph Neural Networks}
Graph Neural Networks (GNNs) \cite{c:gcn,c:garphsage,c:gin,c:gat,c:link_pred,c:predict_n_propogate,c:mp_chem,c:sim_gnn,c:spectural_graph,c:gbp,c:graph_saint,c:dagnn,a:mgnn, a:haqjsk,a:peg} have demonstrated tremendous success in various real-world applications, e.g., social recommendation \cite{c:gnnforsocialrecommendation,c:knowgnnsocialrecomend}, molecule discovery \cite{c:graphaf}, etc. 
Inspired by Convolutional Neural Networks (CNN) \cite{a:cnn}, graph convolution is applied in graph data to make the networks more efficient and convenient. 
Over the years, various convolutional GNNs have been proposed, including spectral-based and spatial-based approaches. Graph Convolutional Networks (GCN) later bridge the gap between spectral-based approaches and spatial-based \cite{c:gcn}, and then spatial-based approaches become more popular as they are efficient, flexible and general. For example, GraphSAGE with its proposed sampling and aggregation strategies \cite{c:garphsage} can be used for inductive learning and large-scale graph learning. Graph Attention Networks (GAT) adopt the self-attention mechanism to differentiate the importance of neighbors \cite{c:gat}. Graph Isomorphism Networks (GIN) introduce a pooling architecture so that they have expressive power as Weisfeiler-Lehman test \cite{c:gin}.

\subsection{Explainability of GNN}

Most of the GNN explanation approaches can be categorized into four types: perturbation-based, gradient-based, decomposition, and surrogate approaches \cite{a:exgnntaxsur}.

Our proposed approach is closely aligned with perturbation-based methods, which study the outcome changes w.r.t.~different input perturbations. 
GNNExplainer \cite{c:gnnexplainer} employs a trainable mask to perturb the data in the input space, to maximize the mutual information between perturbed input data and model outcome, to obtain a subgraph explanation that is relevant for the particular prediction. 
Though PGExlainer \cite{c:pgexplainer} shares the same objective with GNNExplainer, it achieved faster inference time by learning a parameterized mapping from the graph representation space to subgraph space. % parameterized by $\omega$.
CF-GNNExplainer generates a counterfactual explanation that can flip the model prediction subject to a minimal perturbation \cite{c:cf_gnnexplainer}. 
RG-Explainer \cite{c:rgexplainer} leverages Reinforcement Learning (RL) algorithm to generate the explanation by sequentially adding nodes (action) based on the current generated explanation (state), which has the similar objective (reward) with \cite{c:pgexplainer}.
RC-Explainer \cite{a:rcexplainer} also employs a RL algorithm to search for the explanation that maximizes the causal effect obtained by the edge perturbation. 
$\text{CF}^2$ \cite{c:cf2},  arguably the most closely related work to our approach, utilizes both factual and counterfactual reasoning to generate heuristic necessary and sufficient explanations. The main difference between $\text{CF}^2$ and ours is that the former searches for a trade-off explanation between necessity and sufficiency, while our work searches for the most necessary and sufficiency explanation. Besides, $\text{CF}^2$ only samples one counterfactual (necessary) for computing the necessary strength, which is a degradation to ours.

Gradient-based approaches approximate the importance of an input using the gradients of its outcome obtained by back-propagation. The saliency map approach, which is used to indicate the input importance, is obtained by computing the squared norm of the gradients \cite{a:techingnn}. In another method called Guided Backpropagation (GBP), the negative gradients are clipped during back-propagation as negative gradients are challenging to explain \cite{a:techingnn}.
 
Decomposition approaches aim to decompose the model outcome into several terms as the importance scores of the corresponding input feature. Layer-wise Relevance Propagation (LRP) decomposes the GNN output into node importance scores, whereas the edge importance scores cannot be provided \cite{a:techingnn}. EBP \cite{c:exmethodsgnn} shares a similar idea with LRP, while it is based on the law of total probability.

Surrogate approaches leverage a simple and interpretable model to approximate the behavior of a complex model locally. GraphLIME \cite{a:graphlime} extends the LIME algorithm \cite{c:lime}, and employs a Hilbert-Schmidt Independence Criterion Lasso as a surrogate to approximate the GNN instance. In particular, the weights of the surrogate model indicate the importance scores of the nodes. In PGM-Explainer \cite{c:pgmexplainer}, a probabilistic graphical model is utilized as surrogate for explaining the GNN instance.

\subsection{Casual Explainability}
Causal inference has a long history in statistics \cite{b:causality} and there is now an increasing interest in solving crucial problems of machine learning that benefit from causality \cite{avx:causalsurvey,c:causalvae,c:invrep,c:bandconf}, including explainability. The well-known perturbation-based approaches such as LIME \cite{c:lime}, Shapley values \cite{c:shap} implicitly use causal inference to estimate the attribution scores, which can be viewed as a special case of causal effect. Besides, \citet{c:att_causal} views the neural network architecture as a Structural Causal Model (SCM) and estimates the average causal effect upon it. 
Also, there are a few works \cite{c:pns_xai,c:ex_pcc,c:causal_rational,c:causal_ex_xai} utilizing the aspects of necessity and sufficiency for explainability via causal interpretations, which are the most related works to ours.

\section{Preliminary}
\subsection{Causality}
Here we introduce the basic causality preliminaries to enhance the understanding of this work. 
The identification of PNS, PN, and PS requires the counterfactual estimation. In the literature of \cite{b:causality}, counterfactual is obtained by intervention under the structural causal model (SCM), which is defined in Definition~\ref{def:scm}.

\begin{definition}
\label{def:scm}
(Structural Causal Model). A structural causal model is a triple:
\begin{equation*}
\small
M=(\mathbf{V},\mathbf{U},F),
\end{equation*}
where 
\begin{enumerate}
    \item $\mathbf{V}$ is a set of variables called \emph{endogenous}, that are determined by variables in the model, i.e., $\mathbf{U}\cup \mathbf{V}$.
    \item $\mathbf{U}$ is a set of variables called \emph{exogenous}, that are determined by factors outside the model.
    \item $F$ is a set of functions where each $f_i$ is a mapping from $(\mathbf{V}\setminus \mathbf{v}_i)\times \mathbf{U}$ to $\mathbf{v}_i$, i.e.,
    \begin{equation*}
    \small
    \mathbf{v}_i = f_i(Pa(\mathbf{v}_i),\mathbf{u}_i),
    \end{equation*}
    where $Pa(\mathbf{v}_i)$ are the parent variables of $\mathbf{v}_i$, and $\mathbf{u}_i$ is the exogenous of $\mathbf{v}_i$.
\end{enumerate}
\end{definition}
The SCM can be associated with a directed acyclic graph (DAG), where each node corresponds to variables in $V$ and the directed edges point from members of $Pa(\mathbf{v}_i)$ toward $\mathbf{v}_i$.
A SCM can properly model the data-generating process through functional mechanisms, that is, an endogenous $\mathbf{V}_i$ is determined by its parents $Pa(\mathbf{v}_i)$ and exogenous $\mathbf{u}_i$ via the function $f_i$, denoted as $\mathbf{v}_i = f_i(Pa(\mathbf{v}_i),\mathbf{u}_i)$. The functional characterization in SCM provides a convenient language for specifying how the resulting distribution would change in response to interventions. 
Regarding intervention, the simplest intervention such as an intervention of $do(\mathbf{v}_i=v_i')$, amounts to removing the old generating mechanism $\mathbf{v}_i=f_i(Pa(\mathbf{v}_i),\mathbf{u}_i)$ from the SCM and substituting $\mathbf{v}_i=v_i'$ in the remaining generating equations.
The concept of counterfactual refers to the consequences of the interventions, given certain facts. In particular, the given certain facts provide evidence about the actual state of the world, which is exogenous in the literature of \cite{b:causality}. The counterfactual is obtained by intervening on some variables under the SCM while keeping the actual state (exogenous) the same.

\subsection{Graph Neural Networks}

Graph neural networks are capable of incorporating both graph structure and node features into representations in an end-to-end fashion, to facilitate downstream tasks such as node classification task and graph classification task. 
In particular, in the $k$-th layer of GNNs, the learning process of the representation of node $v$ can be divided into the following three steps: 
\begin{itemize}
    \item First, obtaining message $\mathbf{m}_{v,u}^{(k)}$ for any node pair $(v,u)$ through a message function $MSG$:
    \begin{equation*}
    \small
        \mathbf{m}_{v,u}^{(k)}=MSG(\mathbf{h}_v^{(k-1)},\mathbf{h}_u^{(k-1)},\mathbf{e}_{v,u}),
    \end{equation*}
    where $\mathbf{h}_v^{(k-1)}$ and $\mathbf{h}_u^{(k-1)}$ denote the representations of nodes $v$ and $u$ in the $(k-1)$-the layer, and $\mathbf{e}_{v,u}$ denotes the entry (relation) between nodes $v$ and $u$.
    \item Second, aggregating messages from node $v$'s neighbors $\mathcal{N}_v$ and calculating the an aggregated message $\mathbf{M}_v^{(k)}$ via a aggregating function $AGG$:
    \begin{equation*}
    \small
        \mathbf{M}_v^{(k)}=AGG(\mathbf{m}_{v,u}^{(k)}|u\in\mathcal{N}_v).
    \end{equation*}
    \item Third, updating node $v$'s representation $\mathbf{h}_v^{(k)}$ using the aggregated messages $\mathbf{M}_v^{(k)}$ and node $v$'s representation in the previous layer $\mathbf{h}_v^{(k-1)}$ via a update function $UPDATE$:
    \begin{equation*}
    \small
        \mathbf{M}_v^{(k)}=UPDATE(\mathbf{M}_v^{(k)},\mathbf{h}_v^{(k-1)}).
    \end{equation*}
\end{itemize}
After obtaining the representation of each node, a node-level read-out and a graph-level read-out can be applied for node classification task and graph classification task respectively.

\subsection{Probability of Necessity and Sufficiency}
As we discussed before, a convincing explanation should be necessary and sufficient. 
To quantify the degree of necessity and sufficiency of an explanation to the model outcome, the Probability of Necessity and Sufficiency (PNS) is formally defined as follows. 
\begin{definition}
\label{def:pns}
(Probability of necessity and sufficiency \cite{b:causality}). 
\begin{equation}
\label{eq:pns}
\small
\text{PNS}(\xi)=P(\mathbf{Y}_{\xi^{c}}\neq\hat{y},\mathbf{Y}_{\xi}=\hat{y}),
\end{equation}
where $\mathbf{Y}_{\xi^{c}}$ and $\mathbf{Y}_{\xi}$ are the potential outcome variables under the treatments $\xi^{c}$ and $\xi$ respectively, and $\xi^{c}$ is the complementary event of $\xi$.
\end{definition}

PNS measures the necessity and sufficiency of treatment $\xi$ to model outcome $\hat{y}$ in probability space. Intuitively, PNS indicates the probability that the outcome $\hat{y}$ responds to both treatments $\xi$ and $\xi^{c}$. However, direct optimization of the objective is intractable, given the non-identifiability of PNS shown in Eq.~\eqref{eq:pns} due to the potential violation of monotonicity as defined in Definition \ref{def:monotonicity}, as well as the challenges of counterfactual estimation \cite{b:causality}.

\begin{definition}
\label{def:monotonicity}
(Monotonicity). The model outcome $\mathbf{Y}$ is monotonic relative to the explanation event $\mathbf{\xi}$ if and only if:
\begin{equation*}
\small
(\mathbf{Y}_{\xi}\neq\hat{y})\wedge(\mathbf{Y}_{\xi^{c}}=\hat{y})=false.
\end{equation*}
\end{definition}

%\xi_{E}, \xi_{N}, \xi_{E, N}, \xi_{E, N}^c

Monotonicity indicates that a change from $\xi^c$ to $\xi$ cannot assure the outcome also changes from $\mathbf{Y}=\hat{y}$ to $\mathbf{Y}\neq\hat{y}$ \cite{a:pns}. 
% Due to the complexity of the neural networks, the assumption of monotonicity may not always hold. 
However, the assumption of monotonicity will not always hold during the explanation searching stage.
Instead of addressing the non-monotonic issue to identify the exact PNS, it is reasonable to maximize a lower bound of $\text{PNS}(\xi)$ as shown in Lemma \ref{lemma:pns_lb} for our objective optimization.

\begin{lemma}
\label{lemma:pns_lb}
The lower bound of $\text{PNS}(\xi)$ is:
\begin{equation}
\label{eq:pns_lb}
\small
\mathop{\max}\{0,P(\mathbf{Y}_{\xi^{c}}\neq\hat{y})+P(\mathbf{Y}_{\xi}=\hat{y})-1\}.
\end{equation}
\end{lemma}

In particular, the lower bound is tight if the assumption of monotonicity holds, as shown in Lemma \ref{lemma:pns_tight}. The proof of Lemma \ref{lemma:pns_tight} is given in \ref{sec:proof_lemma_2}.

\begin{lemma}
\label{lemma:pns_tight}
When the outcome $\mathbf{Y}$ is monotonic relative to explanation event $\xi$, the lower bound in Eq.~\eqref{eq:pns_lb} equals to the exact $\text{PNS}(\xi)$.
\end{lemma}

\section{Methodology}
In this section, we develop our approach NSEG to generate the most \textbf{N}ecessary and \textbf{S}ufficient \textbf{E}xplanations for \textbf{G}NNs by optimizing our objective, the Probability of Necessity and Sufficiency (PNS). We maximize a derived lower bound of PNS since the non-identifiability of PNS and incorporate a proposed SCM of GNN for the identification of the lower bound. Additionally, we introduce a continuous optimization method that utilizes continuous masks and a sampling strategy. This approach enables tractable optimization of large-scale graphs, ensuring the scalability and efficiency of our approach.

\subsection{Problem Definition}
Given a trained GNN model $f_{\theta}$ parameterized by $\theta$ for graph classification task, our task is to explain a specific instance $\mathcal{I}:\hat{y}=f_{\theta}(E_i,X_i)$ produced by the model $f_{\theta}$ by generating a necessary and sufficient explanation in a post-hoc manner, where $\hat{y}$ is the predicted label, $X_i=\{x_v|v\in V_i\}$, and $E_i,V_i$ are edges and vertices of graph $G_i$. 
In our framework, we consider the explanation of a specific instance $\mathcal{I}$ as an event, i.e., $(\mathbf{E}=E_i',\mathbf{X}=X_i')$, that most necessarily and sufficiently causes the model outcome $\mathbf{Y}=\hat{y}$ (with maximum PNS), where $E'_i\subset E_i$ and $X_i'=\{x_v|v\in V_i'\}$ with $V'_i\subset V_i$. Note that our main focus is to identify a set of node features instead of a set of features in the feature dimensions, where the latter has been well investigated in \cite{c:att_causal}.
Similarly, as for the formulation of node classification task, our task is to explain a node-level instance $\mathcal{I}:\hat{y}_i=f_{\theta}^{(i)}(E,X)$ for node $i$ by generating a necessary and sufficient explanation $\mathbf{E}=E',\mathbf{X}=X'$, where $\hat{y}_i$ is the predicted label of node $i$.
Without loss of generality, we formulate our approach in a graph classification fashion in the following paper.
Regarding our notations, we use the bold font notation for random variable to emphasize the distinction between r.v. and its realization.

\subsection{Lower Bound of the Explanation's PNS on GNN}
Although PNS has been formally defined, applying it to GNNs is not a trivial task due to: 1) the combined influence of both structural information (edges) and feature information (node features) on GNN predictions, 2) GNN takes continuous inputs rather than a binary variable $\xi$.
To overcome these challenges, we extend PNS to the graph domain by defining a joint event $(\mathbf{E}=E_i',\mathbf{X}=X_i')$ as the explanation, and its complement event $(\mathbf{E}=E_i',\mathbf{X}=X_i')^c$ which can be derived as $(\mathbf{E}\neq E_i',\mathbf{X}\neq X_i')\vee(\mathbf{E}\neq E_i',\mathbf{X}=X_i')\vee(\mathbf{E}=E_i',\mathbf{X}\neq X_i')$.
Hence, our objective becomes maximizing the probability of necessity and sufficiency w.r.t. $E_i'$ and $X_i'$ as:
\begin{equation}
\label{eq:objective_edge_node}
\small
\mathop{\max}\limits_{E_i',X_i'}{\text{PNS}^{e,f}(E_i',X_i')},
\end{equation}
where $\text{PNS}^{e,f}(E_i',X_i')$ is defined as:
\begin{equation}
\label{eq:pns_gnn}
\small
\begin{aligned}
&\text{PNS}^{e,f}(E_i',X_i')\\
=&P_{\theta}(\mathbf{Y}_{(\mathbf{E}=E_i',\mathbf{X}=X_i')^{c}}\neq\hat{y},\mathbf{Y}_{\mathbf{E}=E_i',\mathbf{X}=X_i'}=\hat{y}),
\end{aligned}
\end{equation}
where $\theta$ denotes the model parameter. The lower bound of $\text{PNS}^{e,f}(E_i',X_i')$ is:
\begin{equation}
\label{eq:pns_lb_ef}
\small
\begin{aligned}
\mathop{\max}\{0,P_{\theta}(\mathbf{Y}_{(\mathbf{E}=E_i',\mathbf{X}=X_i')^{c}}\neq\hat{y})\\
+P_{\theta}(\mathbf{Y}_{\mathbf{E}=E_i',\mathbf{X}=X_i'}=\hat{y})-1\}.
\end{aligned}
\end{equation}

Note that our approach to optimize the above lower bound in Eq.~\eqref{eq:pns_lb_ef} provides a joint explanation of both the edge and node feature. Single edge explanation and single node feature explanation are special cases of our joint explanation. Specifically, the formulations of single-edge and single-node features are familiar with the joint formulation, which are given by:
\begin{equation*}
\label{eq:pns_e_or_f_def}
\small
\begin{aligned}
\text{PNS}^{e}(E_i',X_i)=P_{\theta}(\mathbf{Y}_{\mathbf{E}\neq E_i'}\neq\hat{y},\mathbf{Y}_{\mathbf{E}=E_i'}=\hat{y}|\mathbf{X}=X_i),\\
\text{PNS}^{f}(E_i,X_i')=P_{\theta}(\mathbf{Y}_{\mathbf{X}\neq X_i'}\neq\hat{y},\mathbf{Y}_{\mathbf{X}=X_i'}=\hat{y}|\mathbf{E}=E_i),
\end{aligned}
\end{equation*}
where the lower bound of each is shown as follows respectively.
\begin{equation*}
\label{eq:pns_lb_e_or_f}
\small
\begin{aligned}
\mathop{\max}\{0,P_{\theta}(\mathbf{Y}_{\mathbf{E}=E_i'}\neq\hat{y}|X_i)+P_{\theta}(\mathbf{Y}_{\mathbf{E}=E_i'}=\hat{y}|X_i)-1\},\\
\mathop{\max}\{0,P_{\theta}(\mathbf{Y}_{\mathbf{X}=X_i'}\neq\hat{y}|E_i)+P_{\theta}(\mathbf{Y}_{\mathbf{X}=X_i'}=\hat{y}|E_i)-1\}.
\end{aligned}
\end{equation*}

\subsection{Estimating the Lower Bound via Counterfactual Estimation}

\begin{figure*}
    \centering
    \includegraphics[scale=0.4]{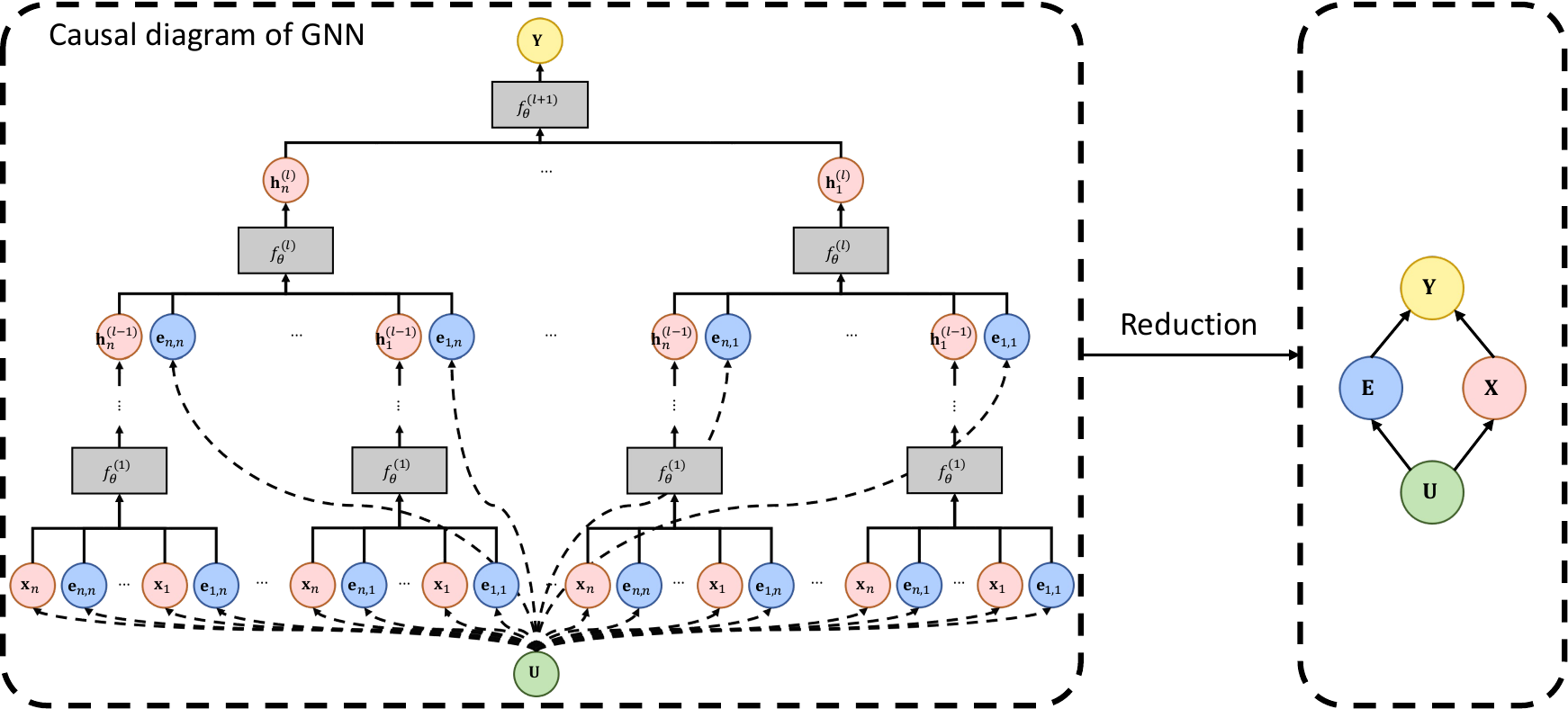}
    \caption{The causal diagram of corresponding SCM of GNN. $\mathbf{h}_{v}^{k}$ denotes the hidden representation of node $v$ in $k$-th layer, $\mathbf{e}_{v,u}$ denotes the entry between $v$ and $u$, $\mathbf{x}_{v}$ denotes the node feature of node $v$, $\mathbf{H}^{(k)}=\{\mathbf{h}_{v}^{(k)}|v\in V\}$ denotes a set of node representations in $k$-th layer, $\mathbf{X}=\{\mathbf{x}_{v}|v\in V\}$ denotes a set of node features, $\mathbf{E}=\{\mathbf{e}_{v,u}|v,u\in V\}$ denotes a set of graph entries.} 
    \label{fig:gnn_scm}
\end{figure*}

\subsubsection{GNN as Structural Causal Model}
Obtaining the probability term in Eq.~\eqref{eq:pns_lb_ef} requires generating counterfactual by intervention under a specific structural causal model (SCM) \cite{b:causality}. In terms of counterfactual, it answers what would the outcome $\mathbf{Y}$ be if $(\mathbf{E}=E_i',\mathbf{X}=X_i')^{c}$ or $(\mathbf{E}=E_i',\mathbf{X}=X_i')$. 
Building upon the interpretation of feed-forward neural networks as SCM for investigating causal attribution \cite{c:causal_explain_s2s,c:att_causal}, we extend the SCM interpretation to GNNs.
In particular, GNNs can be interpreted as directed acyclic graphs with directed edges from the lower layer to the layer above. 
\begin{proposition}
\label{prop:scm_gnn}
An $(l+1)$-layer GNN corresponds to an SCM $M(\{\mathbf{X},\mathbf{E},\mathbf{H}^{(1)},...,\mathbf{H}^{(l)},\mathbf{Y}\}, \allowbreak \mathbf{U}, \{f^{(0)},f_{\theta}^{(1)},...,f_{\theta}^{(l+1)}\})$, where $\mathbf{H}^{(k)}$ denotes a set of node hidden representation after the $k$-th graph convolution layer, i.e., $\mathbf{H}^{(k)}=f_{\theta}^{(k)}(\mathbf{E},\mathbf{H}^{(k-1)})$. $f_{\theta}^{(l+1)}$ denotes the read-out layer, which can be a node-level read-out for node classification tasks or a graph-level read-out for graph classification tasks, with $\mathbf{Y}=f_{\theta}^{(l+1)}(\mathbf{H}^{(l)})$. $\mathbf{U}$ refers to a set of exogenous which act as causal factors for $\mathbf{X}$ and $\mathbf{E}$, i.e., $\mathbf{E},\mathbf{X}=f^{(0)}(\mathbf{U})$.
\end{proposition}

The proof of Proposition~\ref{prop:scm_gnn} is provided in \ref{sec:proof_scm_gnn}. For a more intuitive understanding, the left side of Figure~\ref{fig:gnn_scm} illustrates $M$'s corresponding causal graph. In the $k$-th layer of GNN, the node $v$'s representation $\mathbf{h}_{v}^{(k)}$ is determined by all nodes' representations $\mathbf{H}^{(k-1)}$ in $(k-1)$-th layer and the entry of the graph $\mathbf{E}$, specifically, aggregating node $u$'s representation $\mathbf{h}_{u}^{(k-1)}$ to obtain $\mathbf{h}_{v}^{(k)}$ if the entry $\mathbf{e}_{v,u}$ is not zero. Since our focus is on the mapping from the input to output rather than the full intrinsic mappings in the hidden layer, the SCM of GNN can be reduced to SCM $M(\{\mathbf{E},\mathbf{X},\mathbf{Y}\},\mathbf{U},\{f^{(0)},f_{\theta}\})$ by marginalizing the hidden representations.

\begin{proposition}
\label{prop:scm_gnn_reduced}
The SCM of an $(l+1)$-layer GNN, $M({\{\mathbf{X},\mathbf{E},\mathbf{H}^{(1)},...,\mathbf{H}^{(l)},\mathbf{Y}\}}, \allowbreak {\mathbf{U}}, {\{f^{(0)}, f_{\theta}^{(1)},...,f_{\theta}^{(l+1)}\}})$ can be reduced to $M(\{\mathbf{E},\mathbf{X},\mathbf{Y}\},\mathbf{U},\{f^{(0)},f_{\theta}\})$.
\end{proposition}

The proof of Proposition~\ref{prop:scm_gnn_reduced} is presented in \ref{sec:proof_scm_gnn_reduced}. Intuitively, marginalizing the hidden representations is analogous to deleting edges connecting the hidden representations and creating new directed edges from the parents of the deleted to their respective child vertices in the causal diagram depicted on the left side of Figure~\ref{fig:gnn_scm}. The corresponding causal diagram of the reduced SCM in Proposition~\ref{prop:scm_gnn_reduced} is shown on the right side of Figure~\ref{fig:gnn_scm}.

\subsubsection{Counterfactual Estimation}
Given the reduced SCM depicted in Proposition~\ref{prop:scm_gnn_reduced}, we can obtain the interventional probability via $do$-calculus, by controlling $\mathbf{E}$ and $\mathbf{X}$ for edges and node features respectively.
To facilitate understanding, we present the formulations separately for edges and node features before combining them to give the joint formulation.

\mypara{Counterfactual for Edges}
Regarding edges, the interventional probability of the counterfactual $(\mathbf{E}=E_i')$ can be obtained by intervening on $\mathbf{E}$, i.e., replacing the causal mechanism from exogenous $\mathbf{U}$ to the edge $\mathbf{E}$ with the intervention $do(\mathbf{E}=E_i')$ while keeping other mechanisms unperturbed, i.e., $\mathbf{X}=X_i$:
\begin{equation}
\label{eq:do_e_eq}
\small
\begin{aligned}
&P_{\theta}(\mathbf{Y}_{\mathbf{E}=E_i'}=\hat{y}|\mathbf{X}=X_i)\\
=&P_{\theta}(\mathbf{Y}=\hat{y}|do({\mathbf{E}=E_i'}),\mathbf{X}=X_i)\\
=&f_{\theta}^{\hat{y}}(E_i',X_i),
\end{aligned}
\end{equation}
where $f_{\theta}^{\hat{y}}(.,.)$ outputs the probability of the class $\hat{y}$ and $\hat{y}$ is the label predicted by the model. \yxzhu{Note that we only alter the generating mechanism $f^{(0)}$ for $\mathbf{E}$ and the GNN model $f_{\theta}$ remains unchanged.} As for the interventional probability of the counterfactual $(\mathbf{E}\neq E_i')$, 
\begin{equation}
\label{eq:do_e_neq}
\small
\begin{aligned}
&P_{\theta}(\mathbf{Y}_{\mathbf{E}\neq E_i'}\neq\hat{y}|\mathbf{X}=X_i)\\
=&P_{\theta}(\mathbf{Y}\neq\hat{y}|do({\mathbf{E}\neq E_i'}),\mathbf{X}=X_i)\\
=&P_{\theta}(\mathbf{Y}\neq\hat{y}|{\mathbf{E}\in\mathcal{E}_i\setminus E_i',\mathbf{X}=X_i})\\
=&1-\mathbb{E}_{\overline{E}_{i}'}[f_{\theta}^{\hat{y}}(\overline{E}_{i}',X_i)]
,
\end{aligned}
\end{equation}
where $\overline{E}_{i}'\sim p(\overline{E}_{i}'|{\mathcal{E}_i\setminus E_i'})$ which can be specified by $p(\mathbf{U})$ according to the corresponding SCM, and $\mathcal{E}_i$ is the sub-edge space of graph $G_i$. 
In the absence of any prior knowledge of $p(\overline{E}_{i}'|{\mathcal{E}_i\setminus E_i'})$, it is reasonable to assume $p(\overline{E}_{i}'|{\mathcal{E}_i\setminus E_i'})$ is uniformly distributed over the sub-edge space $\mathcal{E}$, which encourage the exploration of all $(\mathbf{E}\neq E_i')$. 
Prior knowledge $p(\mathbf{U})$ can also be incorporated to refine the estimation of counterfactual.

\mypara{Counterfactual for Node Features}
Our goal is to identify a subset of node features that cause the prediction, that is, features in a subset of nodes $V_i'$, instead of identifying a subset of feature in feature dimensions. One way to incorporate sub-node structure with the node features is to set the features outside $V_i'$ to $\mathbf{0}$ while keeping features inside $V_i'$ the same, since the zero-valued features have no impact in feed-forward process. Thus, the interventional probability of counterfactual $(\mathbf{X}=X_i')$ can be obtained by intervening on $\mathbf{X}$, i.e., replacing the causal mechanism from $\mathbf{U}$ to $\mathbf{X}$ with the intervention $do(\mathbf{X}=X_i')$ while keeping other mechanisms unperturbed, which is given by,
\begin{equation}
\label{eq:do_f_eq}
\small
\begin{aligned}
&P_{\theta}(\mathbf{Y}_{\mathbf{X}=X_i'}=\hat{y}|\mathbf{E}=E_i)\\
=&P_{\theta}(\mathbf{Y}=\hat{y}|do({\mathbf{X}=X_i'}),\mathbf{E}=E_i)\\
=&f_{\theta}^{\hat{y}}(E_i,X_i') \\
&with \quad X_i'=\{x_v|v\in V_i'\}\cup\{\mathbf{0}|v\notin V_i'\}.
\end{aligned}
\end{equation}

Likewise, the interventional probability of counterfactual $(\mathbf{X}=X_i')$ is given by
\begin{equation}
\label{eq:do_f_neq}
\small
\begin{aligned}
&P_{\theta}(\mathbf{Y}_{\mathbf{X}\neq X_i'}\neq\hat{y}|\mathbf{E}=E_i)\\
=&P_{\theta}(\mathbf{Y}\neq\hat{y}|do({\mathbf{X}\neq X_i'}),\mathbf{E}=E_i) \\
=&\mathbb{E}_{\overline{X}_{i}'}[P_{\theta}(\mathbf{Y}\neq\hat{y}|\mathbf{X}=\overline{X}_{i}',\mathbf{E}=E_i)]\\
=&1-\mathbb{E}_{\overline{X}_{i}'}[f_{\theta}^{\hat{y}}(E_i,\overline{X}_{i}')]\\
&with \quad \overline{X}_{i}'=\{x_v|v\in \overline{V}_{i}'\}\cup\{\mathbf{0}|v\notin \overline{V}_{i}'\},
\end{aligned}
\end{equation}
where $\overline{X}_{i}'\sim p(\overline{X}_{i}'|\overline{V}_{i}')p(\overline{V}_{i}'|{\mathcal{V}_i\setminus V_i'})=p(\overline{V}_{i}'|{\mathcal{V}_i\setminus V_i'})$,
and $\mathcal{V}_i$ is sub-node space of graph $G_i$. 

% Moreover, instead of perturbing the node features with zero vector $\mathbf{0}$ to softly remove a node from graph, it is reasonable to 

\mypara{Joint Formulation of Edge and Node Feature}
The joint formulation of both edge and node feature explanation in Eq.~\eqref{eq:pns_lb_ef} requires us to generate counterfactual for edge and node feature by intervention on both $\mathbf{E}$ and $\mathbf{X}$. Notably, the event $(\mathbf{E}=E_i',\mathbf{X}=X_i')^{c}$ in Eq.~\eqref{eq:pns_lb_ef} can be divided into three sub-events, $(\mathbf{E}\neq E_i',\mathbf{X}\neq X_i')$, $(\mathbf{E}\neq E_i',\mathbf{X}=X_i')$, and $(\mathbf{E}=E_i',\mathbf{X}\neq X_i')$. According to the interventional probability in Eqs.~\eqref{eq:do_e_eq}, \eqref{eq:do_e_neq}, \eqref{eq:do_f_eq}, and \eqref{eq:do_f_neq}, the lower bound of PNS in Eq.~\eqref{eq:pns_lb_ef} can be derived as follows.
\begin{equation}
\begin{aligned}
\small
\label{eq:pns_lb_ef_op}
\mathop{\max}\{0,-&P_{00}\mathbb{E}_{\overline{E}_{i}',\overline{X}_{i}'}[f_{\theta}^{\hat{y}}(\overline{E}_{i}',\overline{X}_{i}')]\\
-&P_{01}\mathbb{E}_{\overline{E}_{i}'}[f_{\theta}^{\hat{y}}(\overline{E}_{i}',X_i')]\\
-&P_{10}\mathbb{E}_{\overline{X}_{i}'}[f_{\theta}^{\hat{y}}(E_i',\overline{X}_{i}')]\\
+&f_{\theta}^{\hat{y}}(E_i',X_i')\},
\end{aligned}
\end{equation}
with:
\begin{equation}
\label{eq:p000110}
\small
\begin{aligned}
P_{00}&=P(\mathbf{E}\neq E_i',\mathbf{X}\neq X_i'|(\mathbf{E}=E_i',\mathbf{X}=X_i')^{c}),\\
P_{01}&=P(\mathbf{E}\neq E_i',\mathbf{X}=X_i'|(\mathbf{E}=E_i',\mathbf{X}=X_i')^{c}),\\
P_{10}&=P(\mathbf{E}=E_i',\mathbf{X}\neq X_i'|(\mathbf{E}=E_i',\mathbf{X}=X_i')^{c}),\\
\end{aligned}
\end{equation}
% Without any prior knowledge regarding $P_{00}$, $P_{01}$, and $P_{10}$ in Eq.~\eqref{eq:p000110}, it is reasonable to assume $P_{00}=P_{01}=P_{10}=\frac{1}{3}$ for the encouragement of fair exploration of each case, intuitively, ensuring to yield the maximum lower bound of PNS for the three cases simultaneously.
\chen{Note that the values of \(P_{00}\), \(P_{01}\) and \(P_{10}\) need to be specified manually. Since these values are interpretable, their ranges can be flexibly defined based on prior knowledge. In this work, we simply assume all possible events derived from \((E_i, X_i)^c\) are equally likely, setting $P_{00}$$=$$P_{01}$$=$$P_{10}$$=$$1/3$.}

\subsection{Generating the Explanation via Lower Bound Optimization}
\mypara{Continuous mask}
Enumerating all possible $E_i'$ and $X_i'$ for objective optimization in large-scale graphs is infeasible. To enhance the scalability of our approach, we adopt a continuous relaxation approach, as in \cite{c:gnnexplainer}, using continuous masks that allow for optimization through gradient descent.
In particular, we design two masks $M_{e}\in [0,1]^m$ and $M_{f}\in [0,1]^n$ to mask the edges $E_i$ and the node feature $X_i$ to obtain $E_i'$ and $X_i'$ respectively, where $m$ is the number of edges, and $n$ is the number of nodes. Intuitively, $M_e^k=0$ indicates that deleting the $k$-th edge from the full edges $E_i$, while $M_e^k=1$ indicates retaining the $k$-th edge.
\begin{equation*}
\label{eq:masking}
\small
\begin{aligned}
E_i'&=M_{e}\odot E_i,\\
X_{i}'&=M_{f}\odot X_i+(1-M_{f})\odot \mathbf{0}=M_{f}\odot X_i,
\end{aligned}
\end{equation*}
where $\odot$ denotes the Hadamard multiplication, and $\mathbf{0}\in 0^{n\times d}$ denotes a zero matrix. After masking, the term $f_{\theta}^{\hat{y}}(E_i',X_i')$ in Eq.~\eqref{eq:pns_lb_ef_op} can be derived as follows.
\begin{equation}
\label{eq:do_ef_11_final}
\small
f_{\theta}^{\hat{y}}(E_i',X_i')=f_{\theta}^{\hat{y}}(M_{e}\odot E_i,M_{f}\odot X_i),
\end{equation}

\mypara{Sampling strategy}
Incorporating the masks to generate samples from $p(\overline{E}_{i}'|{\mathcal{E}_i\setminus E_i'})$ and $p(\overline{V}_{i}'|{\mathcal{V}_i\setminus V_i'})$ in Eq.~\eqref{eq:pns_lb_ef_op}, a heuristic sampling strategy is proposed such that the Monte Carlo estimation of the expectations in Eq.~\eqref{eq:pns_lb_ef_op} are differentiable w.r.t. $M_e$ and $M_f$. Inspired by the reparameterization trick proposed in \cite{c:vae}, an auxiliary variable $\epsilon$ is used in our sampling strategy for the sample generation.

Specifically, the generating process of edge sample $\overline{E}_{i}'$ from $p(\overline{E}_{i}'|{\mathcal{E}_i\setminus E_i'})$ can be expressed as a deterministic function with auxiliary variable $\epsilon_e$, which is:
\begin{equation}
\label{eq:st_e}
\small
\overline{E}_{i}'=(1-M_{e}+\epsilon_{e})\odot E_i, \quad\epsilon_e\sim p(\epsilon_e).
\end{equation}

Intuitively, the term $1-M_{e}$ aims to satisfy the given condition of $\mathbf{E}\in\mathcal{E}_i\setminus E_i'$, and a positive value of $\epsilon_e^k$ increases the weight of the existence of the $k$-th edge, while a negative value decreases the weight. Without any prior knowledge of $p(\overline{E}_{i}'|{\mathcal{E}_i\setminus E_i'})$, it is reasonable to assume $\epsilon_e$ is uniformly distributed to encourage the fair exploration. 
Similarly, the generating process of node feature sample $\overline{X}_{i}'$ from $p(\overline{V}_{i}'|{\mathcal{V}_i\setminus V_i'})$ can be expressed as follows.
\begin{equation}
\label{eq:st_f}
\small
\begin{aligned}
\overline{X}_{i}'=(1-M_{f}+\epsilon_{f})\odot X_i,\quad\epsilon_{f}\sim p(\epsilon_{f}).
\end{aligned}
\end{equation}
Thus, combining with Eqs.~\eqref{eq:st_e} and \eqref{eq:st_f} the term $\mathbb{E}_{\overline{E}_{i}',\overline{X}_{i}'}[f_{\theta}^{\hat{y}}(\overline{E}_{i}',\overline{X}_{i}')]$ in Eq.~\eqref{eq:pns_lb_ef_op} can be derived as:
\begin{equation}
\label{eq:do_ef_00_final}
\small
\begin{aligned}
\mathbb{E}_{\epsilon_e,\epsilon_f}[f_{\theta}^{\hat{y}}((1-M_{e}+\epsilon_{e})\odot E_i,(1-M_{f}+\epsilon_{f})\odot X_i)],
\end{aligned}
\end{equation}
similarly, the term $\mathbb{E}_{\overline{E}_{i}'}[f_{\theta}^{\hat{y}} (\overline{E}_{i}',X_i')]$ is given by:
\begin{equation}
\label{eq:do_ef_01_final}
\small
\begin{aligned}
\mathbb{E}_{\epsilon_e}[f_{\theta}^{\hat{y}}((1-M_{e}+\epsilon_{e})\odot E_i,M_{f}\odot X_i)],
\end{aligned}
\end{equation}
also, the term $\mathbb{E}_{\overline{X}_{i}'}[f_{\theta}^{\hat{y}}(E_i',\overline{X}_{i}')]$ is given by:
\begin{equation}
\label{eq:do_ef_10_final}
\small
\begin{aligned}
\mathbb{E}_{\epsilon_f}[f_{\theta}^{\hat{y}}(M_{e}\odot E_i,(1-M_{f}+\epsilon_{f})\odot X_i)].
\end{aligned}
\end{equation}

Overall, the final optimizable lower bound of PNS combined with continuous masks, PNS$_{lb}^{e,f}$, is given as follows.

% \begin{small}
\begin{equation}
\label{eq:pns_lb_ef_op_final}
\small
\begin{aligned}
\mathop{\max}\{0,-&P_{00}\mathbb{E}_{\epsilon_e,\epsilon_f}[f_{\theta}^{\hat{y}}((1-M_{e}+\epsilon_{e})\odot E_i,(1-M_{f}+\epsilon_{f})\odot X_i)]\\
-&P_{01}\mathbb{E}_{\epsilon_e}[f_{\theta}^{\hat{y}}((1-M_{e}+\epsilon_{e})\odot E_i,M_{f}\odot X_i)]\\
-&P_{10}\mathbb{E}_{\epsilon_f}[f_{\theta}^{\hat{y}}(M_{e}\odot E_i,(1-M_{f}+\epsilon_{f})\odot X_i)]\\
+&f_{\theta}^{\hat{y}}(M_{e}\odot E_i,M_{f}\odot X_i)\},
\end{aligned}
\end{equation}
% \end{small}

Despite joint explanation, for single edge explanation, the final objective is $\mathop{\max}\{0,\mathbb{E}_{\epsilon_e}[-f_{\theta}^{\hat{y}}((1-M_{e}+\epsilon_{e})\odot E_i,X_i)]+f_{\theta}^{\hat{y}}(M_{e}\odot E_i,X_i)\}$. Similarly for single node feature explanation, the final objective is $\mathop{\max}\{0,\mathbb{E}_{\epsilon_f}[f_{\theta}^{\hat{y}}(E_i,(1-M_{f}+\epsilon_{f})\odot X_i)]+f_{\theta}^{\hat{y}}(E_i,M_{f}\odot X_i)\}$.

\subsection{Model Summary}
\begin{figure}[t]
    \centering
    \includegraphics[scale=0.5]{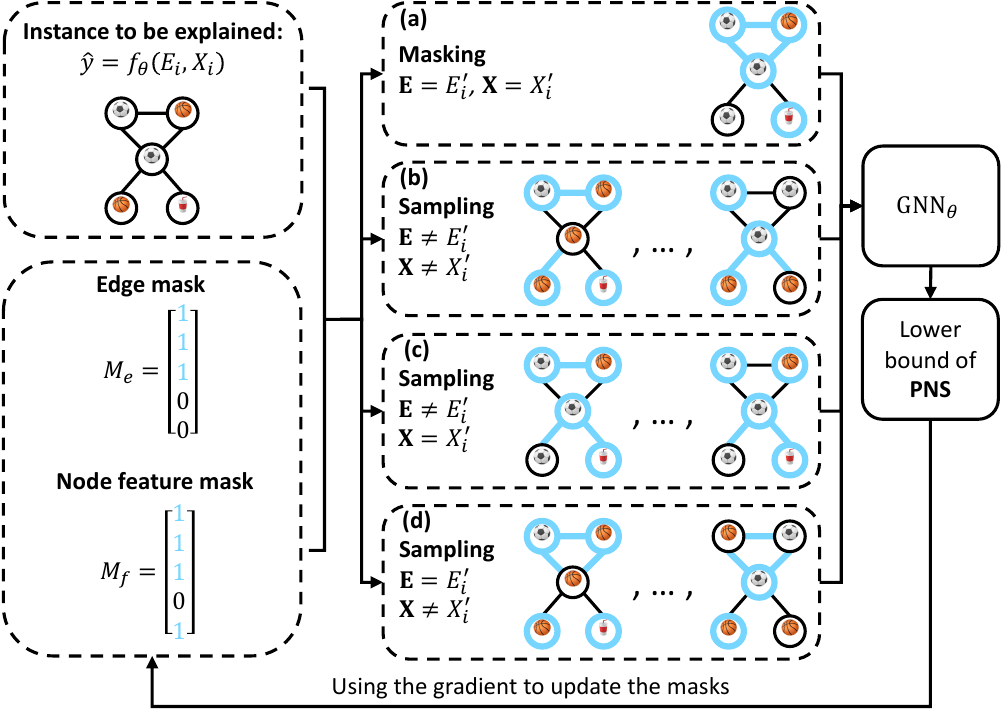}
    \caption{Illustration of the overall framework of NSEG. (a) is the process to obtain Eq.~\eqref{eq:do_ef_11_final}; (b), (c), and (d) are the processes to obtain Eqs.~\eqref{eq:do_ef_00_final}, \eqref{eq:do_ef_01_final}, and \eqref{eq:do_ef_10_final}, respectively.}
    \label{fig:framework}
\end{figure}
The overall framework of NSEG is depicted in Figure~\ref{fig:framework}. To explain an instance $\mathcal{I}:\hat{y}=f_{\theta}(E_i,X_i)$, an edge mask $M_e$ and a node feature mask $M_f$ are used to obtain and maximize the lower bound of PNS in Eq.~\eqref{eq:pns_lb_ef_op_final}. Then the gradients of the overall loss w.r.t. to the masks are employed to update the masks and obtain the necessary and sufficient explanation.
Specifically, we utilize a \emph{mask size} regularization term to enhance the optimization, which is also adopted in prior works \cite{c:gnnexplainer, c:pgexplainer}. Intuitively, the regularization operates as an $L_1$ penalization, compelling the explanation to concentrate on the most important portions of the input. Besides, a \emph{mask entropy} regularization term is added to discretize the mask, i.e., the values of the mask are concentrated around a few scalars when the \emph{mask entropy} is low \cite{a:graphframeex}.
Formally, the overall loss $\mathcal{L}$ is:
\begin{equation}
\label{eq:overall_loss}
\small
\mathcal{L}=-\text{PNS}_{lb}+\alpha_e||M_e||_1+\beta_e Ent(M_e)+\alpha_f||M_f||_1+\beta_f Ent(M_f),
\end{equation}
where $\text{PNS}_{lb}$ is the lower bound of PNS shown in Eq.~\eqref{eq:pns_lb_ef_op_final}, and $Ent()$ is the element-wise entropy to encourage the discretization of the mask.

Formally, the NSEG algorithm is outlined in Algorithm~\ref{alg:nseg}. After obtaining the final masks $M_e$ and $M_f$, the explanation $(E_i',X_i')$ can be obtained via $ extract\_explanation$ in Algorithm~\ref{alg:nseg}. The possible choices for $ extract\_explanation$ can be, extracting the top-K explanation based on the weights of the mask \cite{c:gnnexplainer}, or employing a pre-defined threshold \cite{c:cf2}.

\begin{algorithm}
\caption{The NSEG algorithm}
\label{alg:nseg}
\begin{algorithmic}[1]
\Require The trained GNN model $f_{\theta}$, the instance to be explained $\mathcal{I}:\hat{y}=f_{\theta}(E_i,X_i)$, the hyper-parameters $(\alpha_e,\beta_e,\alpha_f,\beta_f)$, lr $\gamma$, and the \# epochs $n$.
\Ensure The explanation $(E_i',X_i')$.
\State Randomly initialize the $M_e$ and $M_f$
\For{$j=1,...,n$}
\State $\text{PNS}_{\text{lb}}(M_e^{(j)}, M_f^{(j)}) \gets$ the lower bound of PNS calculated by Eq.\eqref{eq:pns_lb_ef_op_final}
\State $\mathcal{L}(M_e^{(j)}, M_f^{(j)}) \gets$ the overall loss calculated by Eq.~\eqref{eq:overall_loss}
\State $(M_e^{(j)}, M_f^{(j)}) \gets$ Update the masks via $\nabla_{M_e,M_f} \mathcal{L}(M_e^{(j-1)}, M_f^{(j-1)})$
\EndFor
\State $(M_e, M_f) \gets (M_e^{(n)}, M_f^{(n)}) $
\State $(E_i',X_i')\gets extract\_explanation(M_e,M_f)$
\end{algorithmic}
\end{algorithm}

\section{NSEG Variants: Optimizing Only PN or PS}

In this section, we present two variants of NSEG(PNS): NSEG(PN) and NSEG(PS). NSEG(PN) focuses on optimizing the Probability of Necessity (PN), considering only the necessity of the explanation. On the other hand, NSEG(PS) concentrates on optimizing the Probability of Sufficiency (PS), considering only the sufficiency of the explanation. These two variants are utilized in our ablation study.

\label{sec:app_pn_ps}
\subsection{PN and PS}
In the NSEG variants, NSEG(PN) and NSEG(PS), our goal is to generate the most necessary explanation via NSEG(PN) and the most sufficient explanation via NSEG(PS). Theoretically, the probability of necessity, PN, and the probability of sufficiency, PS, which can quantify the necessity and sufficiency of explanation $(\xi)$ to the model outcome $\hat{y}$ respectively \cite{a:pns}, are defined in Definition \ref{def:pn} and \ref{def:ps}.

\begin{table*}[t]
    \centering
    \caption{Dataset statistics.}
    \scalebox{0.9}{
\begin{tabular}{lcccccc}
\hline
             & BA-Shapes & Tree-Cycles & Tree-Grid & BA2Motif & Mutagenicity & MSRC\_21 \\ \hline
\#graphs     & 1         & 1           & 1         & 1000     & 4337         & 563      \\
\#avg. nodes & 700       & 871         & 1231      & 25000    & 30.32        & 77.52    \\
\#avg. edges & 2055      & 967         & 1705      & 51392    & 30.77        & 198.32   \\
\#classes    & 4         & 2           & 2         & 2        & 2            & 20       \\ \hline
\end{tabular}
    }
    \label{tab:datasets_statistics}
\end{table*}

\begin{definition}
\label{def:pn}
(Probability of necessity).
\begin{equation}
\label{eq:pn}
\small
\begin{aligned}
\text{PN}(\xi)=P(\mathbf{Y}_{\xi^{c}}\neq\hat{y}|\xi,\mathbf{Y}=\hat{y}).
\end{aligned}
\end{equation}
\end{definition}

\begin{definition}
\label{def:ps}
(Probability of sufficiency).
\begin{equation}
\label{eq:ps}
\small
\begin{aligned}
\text{PS}(\xi)=P(\mathbf{Y}_{\xi}=\hat{y}|\xi^{c},\mathbf{Y}\neq\hat{y}).
\end{aligned}
\end{equation}
\end{definition}

PN captures the probability that the model outcome $\hat{y}$ changes with the absence of the event $\xi$, given the fact that the event $\xi$ happens and the model outcome is $\hat{y}$. In similar, PS captures the probability that the model outcome is $\hat{y}$ with the existence of the event $\xi$, given the fact that the event $\xi$ does not happen, and the model outcome is not $\hat{y}$.

Similar to the generalized formulation in Eq.~\eqref{eq:objective_edge_node}, both PN and PS in Eqs.~\eqref{eq:pn} and \eqref{eq:ps} can be generalized to GNN explanation formulations, which are given as follows.
\begin{equation}
\label{eq:pn_gnn}
\small
\begin{aligned}
&\text{PN}^{e,f}(E_i',X_i')\\
=&P_{\theta}(\mathbf{Y}_{(\mathbf{E}=E_i',\mathbf{X}=X_i')^{c}}\neq\hat{y}|\mathbf{E}=E_i',\mathbf{X}=X_i',\mathbf{Y}=\hat{y}),
\end{aligned}
\end{equation}

\begin{equation}
\label{eq:ps_gnn}
\small
\begin{aligned}
&\text{PS}^{e,f}(E_i',X_i')\\
=&P_{\theta}(\mathbf{Y}_{\mathbf{E}=E_i',\mathbf{X}=X_i'}=\hat{y}|(\mathbf{E}=E_i',\mathbf{X}=X_i')^{c},\mathbf{Y}\neq\hat{y}),
\end{aligned}
\end{equation}

For NSEG(PN), to generate the most necessary explanation of GNN, the objective to maximize the PN defined in Eq.~\eqref{eq:pn_gnn}:
\begin{equation}
\label{eq:objective_pn}
\small
\mathop{\max}\limits_{E_i',X_i'}{\text{PN}^{e,f}(E_i',X_i')}.
\end{equation}

Similar for NSEG(PS), to generate the most sufficient explanation, the objective to maximize the PS defined in Eq.~\eqref{eq:ps_gnn}:
\begin{equation}
\label{eq:objective_ps}
\small
\mathop{\max}\limits_{E_i',X_i'}{\text{PS}^{e,f}(E_i',X_i')}.
\end{equation}

\subsection{Optimization of PN and PS}
The both PN and PS in Eqs.~\eqref{eq:pn_gnn} and \eqref{eq:ps_gnn} are formulated as the probability of counterfactual in the literature of \cite{b:causality}. The identification of both PN and PS requires incorporating with SCM for the recovery of the exogenous \cite{b:causality}. 
However, the deterministic nature of GNN renders the given condition in the formulation of PN and PS not guaranteed, resulting in the non-identifiability of these probabilities.
To better incorporate with the probability outputted by GNN, we derive lower bounds of both PN and PS that can be optimized similarly to PNS.

\begin{proposition}
\label{prop:pn_ps_reduced}
Given the SCM $M$ depicted in Proposition~\ref{prop:scm_gnn_reduced}, the lower bounds of $\text{PN}^{e,f}$ and $\text{PS}^{e,f}$ can be derived as:
\begin{equation*}
\small
\begin{aligned}
\text{PN}_{lb}^{e,f}(E_i',X_i')&=\mathop{\max}\{0,\frac{P_{\theta}(\mathbf{Y}\neq\hat{y}|(\mathbf{E}=E_i',\mathbf{X}=X_i')^{c})\!-\!1}{P_{\theta}(\mathbf{Y}=\hat{y}|\mathbf{E}=E_i',\mathbf{X}=X_i')}\!+\!1\},\\
\text{PS}_{lb}^{e,f}(E_i',X_i')&=\mathop{\max}\{0,\frac{P_{\theta}(\mathbf{Y}=\hat{y}|\mathbf{E}=E_i',\mathbf{X}=X_i')-1}{P_{\theta}(\mathbf{Y}\neq\hat{y}|(\mathbf{E}=E_i',\mathbf{X}=X_i')^{c})}+1\}.
\end{aligned}
\end{equation*}
\end{proposition}
For the proof of Proposition~\ref{prop:pn_ps_reduced} and \chen{the optimization details of the PN and PS lower bounds}, please refer to Sections~\ref{sec:app_proof_lb_pnps} \chen{and~\ref{sec:detail_lbd}}.

\section{Empirical Study}

In this section, we quantitatively and qualitatively evaluate our NESG(PNS) on both synthetic and real-world data to tell:
\begin{itemize}
    \item \textbf{(RQ1)} Are the explanations necessary and sufficient?
    \item \textbf{(RQ1)} Are the necessary and sufficient explanations accurate?
\end{itemize}

\subsection{Dataset}
In this subsection, we will introduce the datasets we used in our experiments, and the statistics of the dataset are presented in Table~\ref{tab:datasets_statistics}.

\mypara{Synthetic Datasets}
We follow the graph generation process in \cite{c:gnnexplainer, c:pgexplainer} and adopt four datasets, BA-Shapes, Tree-Cycles, Tree-Grid, and BA2Motif. Each dataset consists of a base graph and a set of motifs, where for node classification, the class label of each node is determined by its role in the motif, and for graph classification, the class label of each graph is determined by the type of motif in the graph. 
For instance, in BA-Shapes, the motifs are the ``house-shaped'' subgraphs, and the class labels of the nodes are ``bottom'', ``middle'', ``top'' and ``outside''. 
For a node instance, its ground-truth explanation is given by all edges in the motif to which it belongs.
Regarding the explaining indices, we follow the settings in \cite{c:pgmexplainer} for BA-Shapes, Tree-Cycles, and Tree-Grid, and the settings in \cite{c:pgexplainer} for BA2Motif.

\mypara{Real-world Datasets}
Two real-world datasets called Mutagenicity and MSRC\_21 \cite{a:tudataset} are used for \emph{graph classification} in our experiment. Mutagenicity contains chemical compounds that belong to two classes: either mutagenic or not. Each compound is a graph, in which each node is an atom and node features are one-hot encodings of the node atom types. We follow \cite{c:pgexplainer,c:gnnexplainer,c:cf2,a:gcexplainer} and treat the amino-group (-NH2) and nitro-group (-NO2) as the ground-truth explanations for the mutagenic compounds. 
MSRC\_21 is derived from MSRC-v2 \cite{url:msrc}, a benchmark dataset in semantic image processing, where each image belongs to one of the 20 classes describing the scene of the image. A graph is constructed based on the semantic segmentation of each image, in which each node is a super-pixel whose feature is the one-hot embedding of the object semantic type. 
We evaluate on 250 graphs for both Mutagenicity and MSRC\_21.

\begin{table*}[t]
    \centering
    \caption{The hyper-parameters setting of NSEG($\text{PNS}^e$) and NSEG($\text{PNS}^{e,f}$) among experimented datasets. The sub-script ($e$) denotes the hyper-parameters for $M_e$, and the sub-script ($f$) denotes the hyper-parameters for $M_f$.}
    \scalebox{0.8}{
\begin{tabular}{lcccccc}
\hline
            & BA-Shapes & Tree-Cycles & Tree-Grid & \multicolumn{1}{l}{BA2Motif} & Mutagenicity & MSRC\_21 \\ \hline
PNS$^{e}$   &           &             &           & \multicolumn{1}{l}{}         &              &          \\ \hline
$\alpha_e$  & 5.0e-3    & 1.0e-2      & 5.0e-2    & 1.0e-2                       & 1.0e-4       & 1.0e-3   \\
$\beta_e$   & 1.0       & 1.0         & 1.0       & 1.0                          & 1.0e-3       & 1.0      \\ \hline
PNS$^{e,f}$ &           &             &           &                              &              &          \\ \hline
$\alpha_e$  & 5.0e-3    & 1.0e-2      & 1.0e-2    & 1.0e-2                       & 1.0e-4       & 5.0e-4   \\
$\beta_e$   & 1.0       & 1.0         & 1.0       & 1.0                          & 1.0       & 1.0      \\
$\alpha_f$  & 5.0e-3    & 1.0e-3      & 1.0e-2    & 1.0e-2                       & 1.0e-4       & 5.0e-4   \\
$\beta_f$   & 1.0       & 1.0         & 1.0       & 1.0                          & 1.0       & 1.0      \\ \hline
\end{tabular}
}

    \label{tab:hyper}
\end{table*}

\subsection{Experimental Setup}

\mypara{GNN Training Setup}
For both node and graph classification tasks, we employ three layers of Graph Convolutional Networks (GCNs) \cite{c:gcn} with ReLU activations. 
For node classification task, we apply node-level read-out by stacking a fully connected classification layer after the last GCN layer.
For graph classification task, a sum-based read-out is used to obtain a graph representation after the last GCN layer, followed by a fully connected classification layer. 
The detailed training setup and results are provided in \ref{sec:gnn_training}.

\mypara{Evaluation Metrics}
We quantitatively evaluate the explanations from two aspects: (1) \textbf{the necessity and sufficiency (RQ1)}; (2) \textbf{the accuracy when the ground truth explanations are available (RQ2)}. Specifically:
\begin{itemize}
\item \textbf{Necessity and Sufficiency:} 
We utilize \emph{Fidelity+} and \emph{Fidelity-} (abbreviated as \emph{Fid+} and \emph{Fid-}) to quantify the necessity and sufficiency of the explanations, respectively \cite{a:graphframeex}. The higher \emph{Fid+}, the more necessary the explanation, on the contrary, the lower \emph{Fid-}, the more sufficient the explanation. 
Additionally, we use the \emph{charact} score, which combines both \emph{Fid+} and \emph{Fid-}, to measure the overall performance on both necessary and sufficient aspects \cite{a:graphframeex}.
The definitions of \emph{Fid+}, \emph{Fid-} and \emph{charact} scores are shown as follows.
\begin{equation*}
\small
\begin{aligned}
\text{\emph{Fid+}}&=\textstyle 1-\frac{1}{N}\sum_{i=1}^{N}{\mathbb{I}(\mathbf{Y}^{1-M_i}=\hat{y}_i)},\\
\text{\emph{Fid-}}&=\textstyle 1-\frac{1}{N}\sum_{i=1}^{N}{\mathbb{I}(\mathbf{Y}^{M_i}=\hat{y}_i)}, \\
\text{\emph{charact}}&=\frac{2\times\text{\emph{Fid+}}\times(1-\text{\emph{Fid-}})}{\text{\emph{Fid+}}+(1-\text{\emph{Fid-}})},
\end{aligned}
\end{equation*}
where $M$ is the explanation mask, and  $\mathbf{Y}^{1-M}=f_{\theta}(E\odot(1-M_e),X\odot(1-M_{f}))$, and $\mathbf{Y}^{M}=f_{\theta}(E\odot M_e,X\odot M_{f})$.
We use \emph{Fid+}$^c$, \emph{Fid-}$^c$ and \emph{charact}$^c$ to denote these scores for continuous mask explanations. Since the discrete nature of graph, we further discretize the explanations mask via threshold, and compute the \emph{Fid+}$^d$, \emph{Fid- }$^d$ and \emph{charact}$^d$ for discrete mask explanations. 
\item \textbf{Accuracy:} 
We use \emph{Recall@K} and \emph{ROC-AUC} to evaluate the \chen{explanation prediction accuracy on all datasets with ground-truths: BA-Shapes, Tree-Cycles, Tree-Grid, BA2Motif, and Mutagenicity.} In particular, \chen{for most methods}, $K$ corresponds to the number of edges in the ground-truth explanations, with values of 6, 6, 12, 5, 15 for these datasets, respectively. \chen{For PGM-Explainer, which generates node-level explanation}, \( K \) is based on the number of nodes in the ground-truths,  with values of 5, 6, 9, 5 and 10 for these datasets, respectively.
\end{itemize}

\mypara{Hyper-parameter Setting}
We apply a grid search to tune the hyper-parameters for our NSEG.
The detailed hyper-parameters settings of NSEG($\text{PNS}^e$) and NSEG($\text{PNS}^{e,f}$) are shown in Table~\ref{tab:hyper}. 
Regarding the baselines, we carefully tune the hyper-parameters based on their respective reported settings.

\mypara{Baselines}
To verify the effectiveness of PNS, we consider three variants of NSEG, named NSEG(PN), NSEG(PS), and NSEG(PNS), which optimize the lower bound of the probability of necessity, the probability of sufficiency, and the probability of necessity and sufficiency respectively. Furthermore, we use a superscript to indicate if edge ($e$) or feature ($f$) is considered in the explanations, e.g., NSEG(PNS$^e$) means only edge explanations are considered by our full PNS model.
We compare our NSEG with the state-of-the-art baselines including the following:
\begin{itemize}
    \item \textbf{Random} \cite{agarwal2023evaluating} generates random edge explanations.
    \item \textbf{GuidedBP} \cite{c:gbp} generates explanation via gradient with negative gradients clipped during back-propagation.

    \item \textbf{GNNExplainer} \cite{c:gnnexplainer} generates explanations by maximizing the mutual information between explanation subgraph and model prediction.
    \item  \textbf{PGExplainer} \cite{c:pgexplainer} generates explanations from parameterized networks whose objective is to maximize the mutual information, similar to GNNExplainer.
    \item \textbf{PGM-Explainer} \cite{c:pgmexplainer}  generates Bayesian networks as explanations upon the perturbation-prediction data to identify significant nodes. 
    \item \textbf{CF-GNNExplainer} \cite{c:cf_gnnexplainer} generates counterfactual explanations capable of flipping the model prediction subject to minimal perturbation.
    \item  \textbf{$\text{CF}^2$} \cite{c:cf2} generates explanations based on factual and counterfactual reasoning.
\end{itemize}

Note that for those baseline approaches that only generate node explanations (Grad, GuidedBP, and PGM-Explainer), we we will not report the \emph{fidelity} since there is no edge explanation obtained, and we only report the \emph{Recall@K}. 
Regarding the implementations of baselines, we adopt their original settings, as detailed in \ref{sec:baselines}. 
Since all baseline approaches generate only edge explanations instead of the joint explanations of edge and node features, for a fair comparison, we mainly compare the results of NSEG(PNS$^{e}$), NSEG(PN$^{e}$), and NSEG(PS$^{e}$) (which only generate edge explanations) with baseline methods in quantitative analysis, and we also showcase the results of NSEG(PNS$^{e,f}$), which generate joint explanations, in both quantitative and qualitative analyses.

\subsection{Are the Explanations Necessary and Sufficient? (RQ1)}
\label{sec:expt:necessary_sufficient}

\begin{table*}[tp]
    \centering
    \caption{\emph{Fid+$^c$}(\%), \emph{Fid-$^c$} (\%) and \emph{charact$^c$} (\%) of the explanations on GCN. Mean and standard deviation are reported. The best result of each metric is bolded.}
    \addtolength{\tabcolsep}{-1.4mm}
\scalebox{0.78}{
\begin{tabular}{lccccccccc}
\hline
\multicolumn{10}{c}{Node Classification}                                                                                                                                                                                                                                                                                                                                                                                                                              \\ \hline
\multicolumn{1}{l|}{}                    & \multicolumn{3}{c|}{BA-Shapes}                                                                                                          & \multicolumn{3}{c|}{Tree-Cycles}                                                                                                        & \multicolumn{3}{c}{Tree-Grid}                                                                                                          \\
\multicolumn{1}{l|}{}                    & \multicolumn{1}{c}{Fid+$^c$ ($\uparrow$)} & \multicolumn{1}{c}{Fid-$^c$ ($\downarrow$)} & \multicolumn{1}{c|}{charact$^c$ ($\uparrow$)} & \multicolumn{1}{c}{Fid+$^c$ ($\uparrow$)} & \multicolumn{1}{c}{Fid-$^c$ ($\downarrow$)} & \multicolumn{1}{c|}{charact$^c$ ($\uparrow$)} & \multicolumn{1}{c}{Fid+$^c$ ($\uparrow$)} & \multicolumn{1}{c}{Fid-$^c$ ($\downarrow$)} & \multicolumn{1}{c}{charact$^c$ ($\uparrow$)} \\ \hline
\multicolumn{1}{l|}{Random(3-hops)}      & 57.25±1.97                                & 58.20±1.12                                  & \multicolumn{1}{c|}{48.30±1.01}               & 77.22±1.82                                & 78.28±1.28                                  & \multicolumn{1}{c|}{33.88±1.55}               & 87.22±0.55                                & 86.28±0.65                                  & 23.71±0.98                                   \\
\multicolumn{1}{l|}{GuidedBP}            & 47.75±0.00                                & 77.00±0.00                                  & \multicolumn{1}{c|}{31.05±0.00}               & 21.67±0.00                                & 85.56±0.00                                  & \multicolumn{1}{c|}{17.33±0.00}               & 94.17±0.00                                & 26.94±0.00                                  & 82.28±0.00                                   \\
\multicolumn{1}{l|}{GNNExplainer}        & 46.95±0.83                                & 72.00±0.00                                  & \multicolumn{1}{c|}{35.08±0.23}               & 77.61±1.34                                & 18.22±0.72                                  & \multicolumn{1}{c|}{79.63±0.81}               & 90.03±0.90                                & 40.58±0.87                                  & 71.58±0.68                                   \\
\multicolumn{1}{l|}{PGExplainer}         & 58.45±0.90                                & 59.45±0.10                                  & \multicolumn{1}{c|}{47.88±0.26}               & 93.61±3.80                                & 95.56±1.24                                  & \multicolumn{1}{c|}{8.46±2.27}                & 88.06±10.12                               & 94.63±1.41                                  & 10.10±2.53                                   \\
\multicolumn{1}{l|}{CFGNNExplainer}      & 33.85±0.25                                & 79.75±0.00                                  & \multicolumn{1}{c|}{25.34±0.07}               & 78.39±0.62                                & 6.11±0.00                                   & \multicolumn{1}{c|}{85.44±0.37}               & 43.83±0.38                                & 59.53±0.57                                  & 42.08±0.44                                   \\
\multicolumn{1}{l|}{CF$^2$}              & 46.60±0.25                                & 72.00±0.00                                  & \multicolumn{1}{c|}{34.98±0.07}               & 78.89±0.00                                & 21.11±0.00                                  & \multicolumn{1}{c|}{78.89±0.00}               & 43.89±0.00                                & 56.11±0.00                                  & 43.89±0.00                                   \\ 
\multicolumn{1}{l|}{NSEG(PN$^e$)}        & \textbf{100.00±0.00}                      & 66.60±0.12                                  & \multicolumn{1}{c|}{50.07±0.14}               & \textbf{100.00±0.00}                      & \textbf{0.00±0.00}                          & \multicolumn{1}{c|}{\textbf{100.00±0.00}}     & 99.31±0.00                                & 11.53±0.00                                  & 93.58±0.00                                   \\
\multicolumn{1}{l|}{NSEG(PS$^e$)}        & 62.75±0.00                                & \textbf{0.00±0.00}                          & \multicolumn{1}{c|}{77.11±0.00}               & 98.89±0.00                                & \textbf{0.00±0.00}                          & \multicolumn{1}{c|}{99.44±0.00}               & 95.50±0.07                                & \textbf{0.00±0.00}                          & 97.70±0.04                                   \\
\multicolumn{1}{l|}{NSEG(PNS$^e$)}       & 99.60±0.12                                & \textbf{0.00±0.00}                          & \multicolumn{1}{c|}{99.80±0.06}               & \textbf{100.00±0.00}                      & \textbf{0.00±0.00}                          & \multicolumn{1}{c|}{\textbf{100.00±0.00}}     & \textbf{100.00±0.00}                      & \textbf{0.00±0.00}                          & \textbf{100.00±0.00}                         \\
\multicolumn{1}{l|}{NSEG(PNS$^{e,f}$)}   & \textbf{100.00±0.00}                      & \textbf{0.00±0.00}                          & \multicolumn{1}{c|}{\textbf{100.00±0.00}}     & \textbf{100.00±0.00}                      & \textbf{0.00±0.00}                          & \multicolumn{1}{c|}{\textbf{100.00±0.00}}     & \textbf{100.00±0.00}                      & \textbf{0.00±0.00}                          & \textbf{100.00±0.00}                         \\ \hline
\multicolumn{10}{c}{Graph Classification}                                                                                                                                                                                                                                                                                                                                                                                                                             \\ \hline
\multicolumn{1}{l|}{}                    & \multicolumn{3}{c|}{BA2Motif}                                                                                                           & \multicolumn{3}{c|}{Mutagenicity}                                                                                                       & \multicolumn{3}{c}{MSRC\_21}                                                                                                           \\
\multicolumn{1}{l|}{}                    & \multicolumn{1}{c}{Fid+$^c$ ($\uparrow$)} & \multicolumn{1}{c}{Fid-$^c$ ($\downarrow$)} & \multicolumn{1}{c|}{charact$^c$ ($\uparrow$)} & Fid+$^c$ ($\uparrow$)                     & Fid-$^c$ ($\downarrow$)                     & \multicolumn{1}{c|}{charact$^c$ ($\uparrow$)} & Fid+$^c$ ($\uparrow$)                     & Fid-$^c$ ($\downarrow$)                     & charact$^c$ ($\uparrow$)                     \\ \hline
\multicolumn{1}{l|}{Random}              & 50.50±0.00                                & 50.50±0.00                                  & \multicolumn{1}{c|}{49.99±0.00}               & 94.64±0.70                                & 94.48±0.30                                  & \multicolumn{1}{c|}{10.43±0.54}                & 80.16±0.78                                & 80.16±0.60                                  & 31.80±0.79                                   \\
\multicolumn{1}{l|}{GuidedBP}            & 50.50±0.00                                & 50.50±0.00                                  & \multicolumn{1}{c|}{50.00±0.00}               & 88.40±0.00                                & 58.80±0.00                                  & \multicolumn{1}{c|}{56.20±0.00}               & 79.60±0.00                                & 77.60±0.00                                  & 34.96±0.00                                   \\
\multicolumn{1}{l|}{GNNExplainer}        & 46.40±1.20                                & 50.50±0.00                                  & \multicolumn{1}{c|}{47.89±0.64}               & 12.80±1.54                                & 89.44±1.42                                  & \multicolumn{1}{c|}{11.52±1.24}               & 87.36±0.54                                & \textbf{2.40±0.00}                          & \textbf{92.20±0.30}                          \\
\multicolumn{1}{l|}{PGExplainer}         & 50.50±0.00                                & 50.50±0.00                                  & \multicolumn{1}{c|}{49.99±0.00}               & 95.20±0.00                                & 95.20±0.00                                  & \multicolumn{1}{c|}{9.14±0.00}                & 80.64±1.78                                & 80.96±2.00                                  & 30.77±2.71                                   \\
\multicolumn{1}{l|}{CFGNNExplainer}      & \textbf{97.50±1.10}                       & 50.50±0.00                                  & \multicolumn{1}{c|}{65.66±0.25}               & 93.76±0.93                                & 95.20±0.00                                  & \multicolumn{1}{c|}{9.13±0.00}                & 76.16±1.49                                & 91.60±0.00                                  & 15.13±0.03                                   \\
\multicolumn{1}{l|}{CF$^2$}              & 50.50±0.00                                & 50.50±0.00                                  & \multicolumn{1}{c|}{49.99±0.00}               & 98.80±0.00                                & 0.80±0.00                                   & \multicolumn{1}{c|}{99.00±0.00}               & \textbf{97.20±0.00}                       & 52.40±0.00                                  & 63.90±0.00                                   \\
\multicolumn{1}{l|}{NSEG(PN$^e$)}        & 81.90±0.92                                & 18.30±0.75                                  & \multicolumn{1}{c|}{81.80±0.58}               & \textbf{99.90±0.17}                       & 83.30±0.17                                  & \multicolumn{1}{c|}{28.62±0.25}               & 96.40±0.00                                & 51.70±0.17                                  & 64.36±0.15                                   \\
\multicolumn{1}{l|}{NSEG(PS$^e$)}        & 0.00±0.00                                 & 50.50±0.00                                  & \multicolumn{1}{c|}{0.00±0.00}                & 91.33±0.50                                & \textbf{0.27±0.19}                          & \multicolumn{1}{c|}{\textbf{95.35±0.21}}      & 43.60±0.33                                & 55.87±0.19                                  & 43.86±0.25                                   \\
\multicolumn{1}{l|}{NSEG(PNS$^e$)}       & 83.50±1.70                                & \textbf{17.60±0.37}                         & \multicolumn{1}{c|}{\textbf{82.94±1.01}}      & 99.60±0.00                                & 0.80±0.00                                   & \multicolumn{1}{c|}{\textbf{99.40±0.00}}      & 96.00±0.00                                & 33.12±0.47                                  & 78.84±0.33                                   \\
\multicolumn{1}{l|}{NSEG(PNS$^{e,f}$)}   & 73.50±1.38                                & 47.90±0.73                                  & \multicolumn{1}{c|}{60.97±0.69}               & 95.12±0.16                                & 86.48±0.78                                  & \multicolumn{1}{c|}{23.67±1.20}               & 93.52±0.39                                & 61.44±0.20                                  & 54.61±0.24                                   \\ \hline
\end{tabular}
}
\label{tab:fidc_all}
\end{table*}

\begin{figure*}[t]
    \centering
    \includegraphics[scale=0.55]{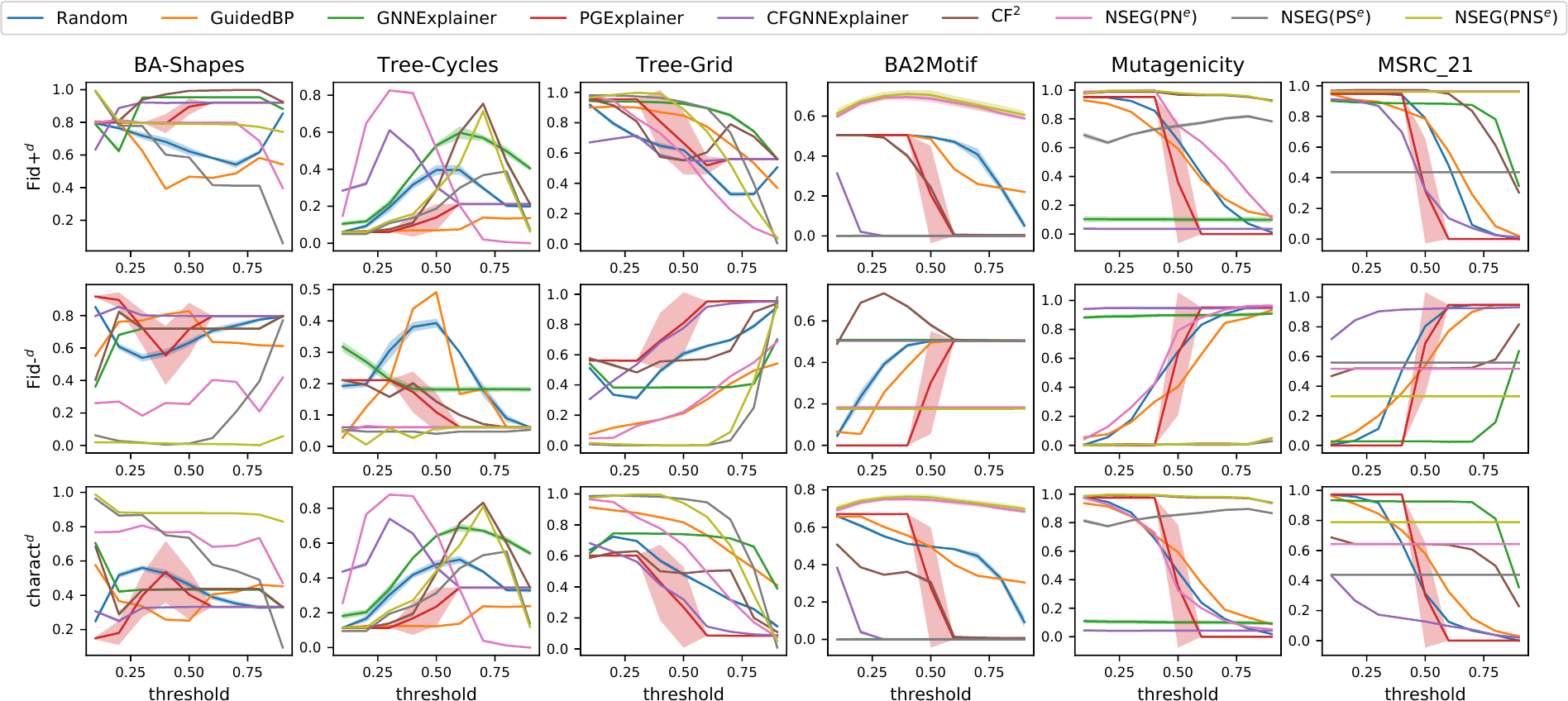}
    \caption{\emph{Fid+}$^d$, \emph{Fid-}$^d$, and \emph{charact}$^d$ w.r.t. threshold on BA-Shapes, Tree-Cycles, Tree-Grid, Mutagenicity, and MSRC\_21 datasets.}
    \label{fig:fid_d}
\end{figure*}

We evaluate the necessity and sufficiency aspects of the explanations based on the \emph{Fid+}, \emph{Fid-}, and \emph{charact} metrics \cite{a:graphframeex} in Table~3 and Figure~4 for continuous and discrete explanations, respectively.

First, when comparing with the baselines, NSEG(PNS) (including NSEG(PNS$^{e}$) and NSEG(PNS$^{e,f}$)) consistently outperforms the other methods, achieving the highest \emph{charact$^c$} scores for continuous explanations in most cases and the second winner on MSRC\_21 dataset. 
When discrete explanations are obtained by thresholding, NSEG(PNS) achieves the best and relatively stable \emph{Fid+$^d$}, \emph{Fid-$^d$}, and \emph{charact$^d$} at most cases and the second best on MSRC\_21.
The second best results on MSRC\_21 can be due to, the winner GNNExplainer that searches for sufficient explanations yields the extremely lowest \emph{Fid-$^c$} and \emph{Fid-$^d$}, thus leads to the highest \emph{charact$^c$} and \emph{charact$^d$}, respectively.

Secondly, among the variants of NSEG, we observe that NSEG(PN$^e$) provides higher \emph{Fid+$^c$} and \emph{Fid+$^d$} scores compared to NSEG(PS$^e$), indicating that NSEG(PN$^e$) focuses more on necessity. Conversely, NSEG(PS$^e$) emphasizes sufficiency, as reflected by its lower \emph{Fid-$^c$} and \emph{Fid-$^d$} scores compared to NSEG(PN$^e$). The full model, NSEG(PNS$^e$), combines the benefits of both aspects, achieving the highest \emph{charact$^c$} and \emph{charact$^d$} scores in most cases, with one exception on \emph{charact$^d$} on Tree-Cycles datasets. This can be due to, the discretization process damages the necessary aspect of the explanation leading to extremely low \emph{Fid+$^d$}, thus causes low \emph{charact$^d$}.

In summary, NSEG(PNS) consistently generates the most necessary and sufficient explanations in most cases compared to the baselines and other variants of our approach.

\subsection{Are the Necessary and Sufficient Explanations Accurate? (RQ2)}
We employ \emph{Recall@K} and \emph{ROC-AUC} to judge if the necessary and sufficient explanations are accurate. As there is only edge-based ground truth on the first three synthetic datasets, we only generate edge explanations using NSEG(PNS$^e$) on those datasets compared with the other baselines, as shown in Table~\ref{tab:comparison}.

The results indicate that our approach generally achieves the highest \emph{Recall@K} except on Tree-Cycles datasets, and slightly lower \emph{ROC-AUC} on Tree-Cycles, BA2Motif, and Mutagenicity. The slightly lower \emph{Recall@K} and \emph{ROC-AUC} of our approach may be attributed to the fact that NSEG prioritizes finding the most necessary and sufficient explanations rather than focusing on ranking the explanations. However, \emph{Recall@K} and \emph{ROC-AUC} metrics are sensitive to ranking, which could explain the relative difference in performance on datasets like Tree-Cycles.

\begin{table*}[t]
    \centering
\caption{\emph{Recall@K} (\%) and \emph{ROC-AUC} (\%) of the explanations on GCN. Mean and standard deviation are reported. The best result of each metric is bolded. Approach with symbol * outputs node-level explanations. \chen{The $K$ values for \emph{Recall@K} in the table are for methods generating edge-level explanations. For methods generating node-level explanations, the $K$ values for these datasets are 5, 6, 9, 5, and 10.}}
\scalebox{0.68}{
\begin{tabular}{l|cc|cc|cc|cc|cc}
\hline
                  & \multicolumn{2}{c|}{BA-Shapes}            & \multicolumn{2}{c|}{Tree-Cycles}          & \multicolumn{2}{c|}{Tree-Grid}            & \multicolumn{2}{c|}{BA2Motif}             & \multicolumn{2}{c}{Mutagenicity} \\
                  & \chen{Recall@6}            & ROC-AUC             & \chen{Recall@6}            & ROC-AUC             & \chen{Recall@12}            & ROC-AUC             & \chen{Recall@5}            & ROC-AUC             & \chen{Recall@15}        & ROC-AUC        \\ \hline
Random            & 33.33±0.38          & 50.09±0.29          & 66.94±0.31          & 50.82±0.38          & 64.04±0.26          & 49.82±0.21          & 22.19±0.97          & 50.45±0.80          & 40.20±0.87      & 49.71±0.85     \\
GuidedBP          & 88.38±0.00          & 99.63±0.00          & 77.69±0.00          & 74.50±0.00          & 72.11±0.00          & 67.63±0.00          & 64.07±0.00          & \textbf{88.44±0.00} & 82.75±0.00      & \textbf{84.06±0.00}     \\
GNNExplainer      & 79.47±0.41          & 79.23±0.44          & 67.16±0.26          & 54.89±0.57          & 68.78±0.03          & 61.59±0.26          & 19.38±0.48          & 50.78±0.38          & 80.43±0.70      & 69.17±0.15     \\
PGExplainer       & 79.89±11.35         & 96.31±4.04          & \textbf{83.61±5.00} & \textbf{79.47±5.00} & 66.43±1.17          & 54.09±6.17          & 49.43±4.30          & 82.59±2.37          & 78.81±6.75      & 79.66±5.19     \\
PGM-Explainer*    & 65.01±0.69          & -                   & 82.35±0.69          & -                   & 72.91±0.20          & -                   & 26.48±1.15          & -                   & 46.55±1.63               & -              \\
CFGNNExplainer    & 75.42±0.07          & 95.40±0.11          & 73.41±0.05          & 66.97±0.13          & 72.52±0.07          & 68.21±0.10          & 60.25±0.18          & 85.52±0.11          & 58.64±0.26      & 59.64±0.07     \\
CF$^2$            & 86.07±0.20          & 99.31±0.00          & 68.71±0.06          & 59.51±0.09          & 69.54±0.08          & 62.56±0.14          & 27.57±0.44          & 66.13±0.24          & 70.58±0.50      & 76.50±0.08     \\
NSEG(PNS$^e$)     & 86.48±0.12          & 98.44±0.08          & 71.92±0.16          & 66.57±0.08          & \textbf{73.27±0.07} & \textbf{69.17±0.06} & \textbf{68.75±0.59} & 85.06±0.23          & 67.95±0.19      & 74.03±0.12     \\
NSEG(PNS$^{e,f}$) & \textbf{89.96±0.07} & \textbf{99.67±0.00} & 72.82±0.13          & 63.13±0.14          & 72.62±0.07          & 67.32±0.05          & 61.43±0.20          & 86.33±0.12          & \textbf{84.69±0.69}      & 70.97±0.37     \\ \hline
\end{tabular}
}
    \label{tab:comparison}
\end{table*}

% \subsection{Influence of $\alpha$}

\begin{figure*}[t]
    \centering
    \includegraphics[scale=0.45]{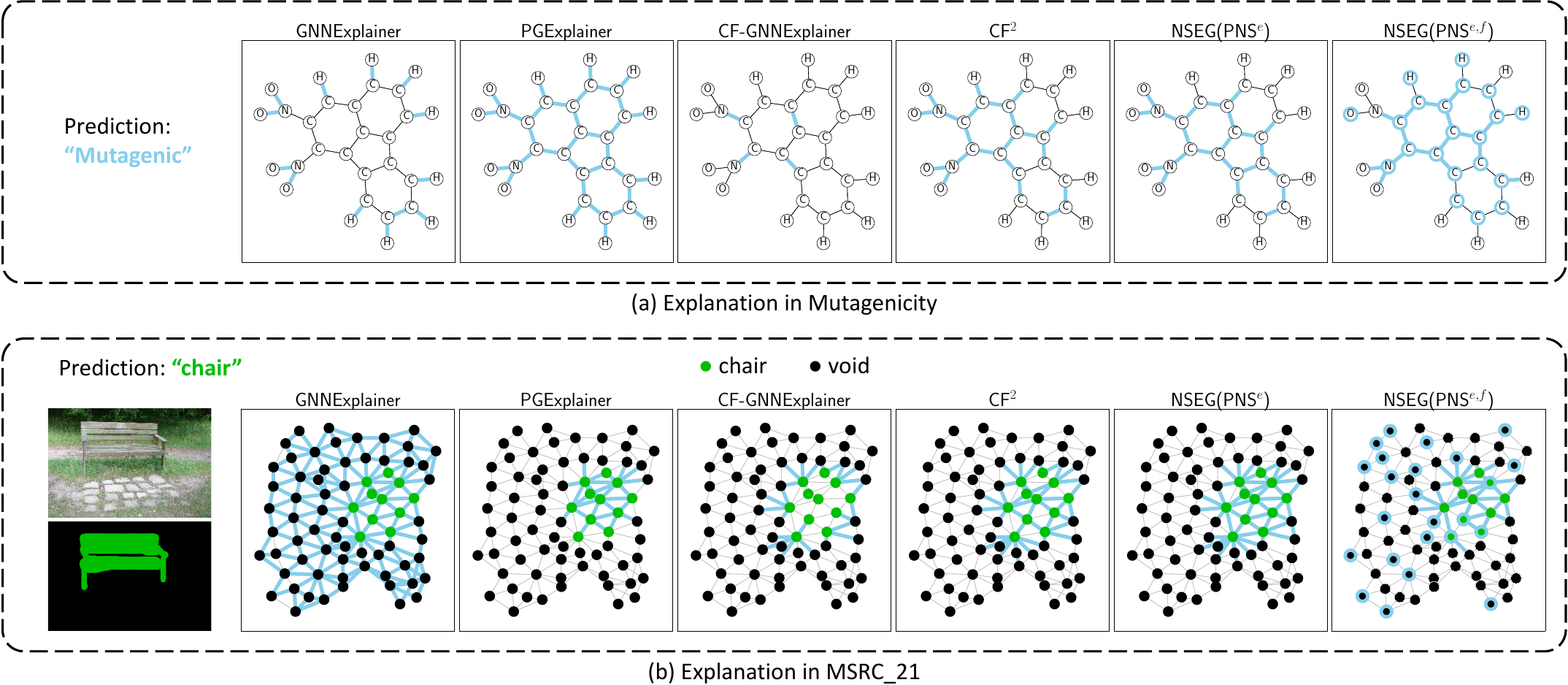}
    \caption{Explanations of GNNExplainer, PGExplainer,CF-GNNExplainer, CF$^2$, NSEG(PNS$^{e}$), and NSEG(PNS$^{e,f}$) in Mutagenicity and MSRC\_21, where the explanations are highlighted in \textcolor{SkyBlue}{blue}.}
    \label{fig:qualitative_baselines}
\end{figure*}

\begin{figure}[h]
    \centering
    \includegraphics[scale=0.15]{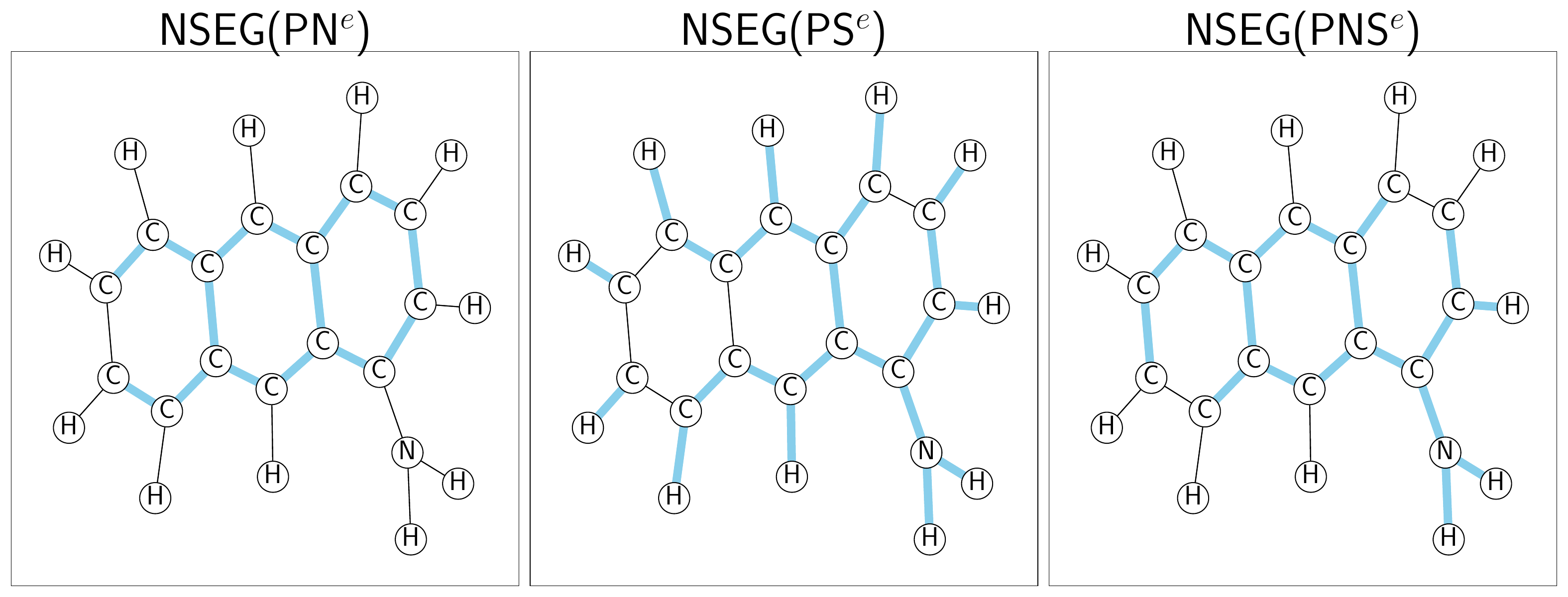}
    \caption{Explanations of NSEG variants, NSEG(PN$^{e}$), NSEG(PS$^{e}$), and NSEG(PNS$^{e}$) on a Mutagenicity instance, where the explanations are highlighted in \textcolor{SkyBlue}{blue}.}
    \label{fig:qualitative_variants}
\end{figure}
\begin{figure}[h]
    \centering
    \subfigure[Explanation of an incorrectly predicted instance. The GNN falsely pays more attention to the nodes ``sky'', ``road'', and ``building'' and the edges among them.]{
    \label{subfig:case_a}
    \includegraphics[scale=0.45]{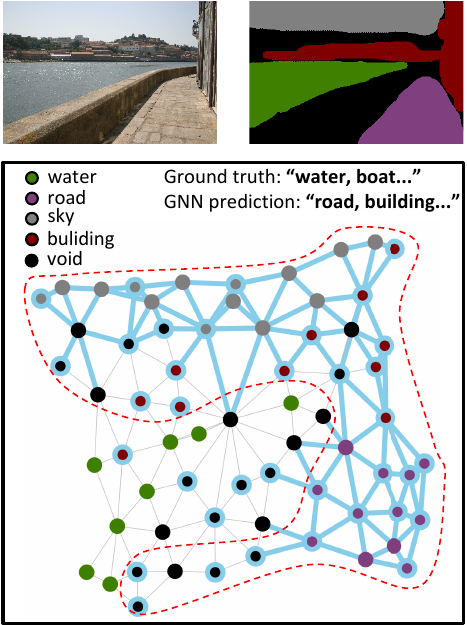}
    }
    \hspace{2mm}
    \subfigure[Explanation of a correctly predicted instance. The GNN correctly pays more attention to the nodes ``dog'' and the edges among them.]{
    \label{subfig:case_b}
    \includegraphics[scale=0.45]{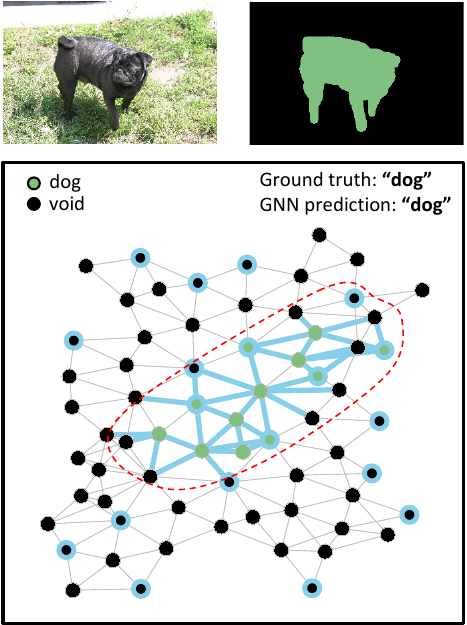}
    }
    \vspace{-2.5mm}
    \caption{Explanations of NSEG(PNS$^{e,f}$) on two MSRC\_21 instances, where the explanations are highlighted in \textcolor{SkyBlue}{blue} and enclosed by a \textcolor{red}{red} dashed line.} 
    \label{fig:case_study}
\end{figure}

\begin{figure*}[h]
    \centering
    \includegraphics[scale=0.18]{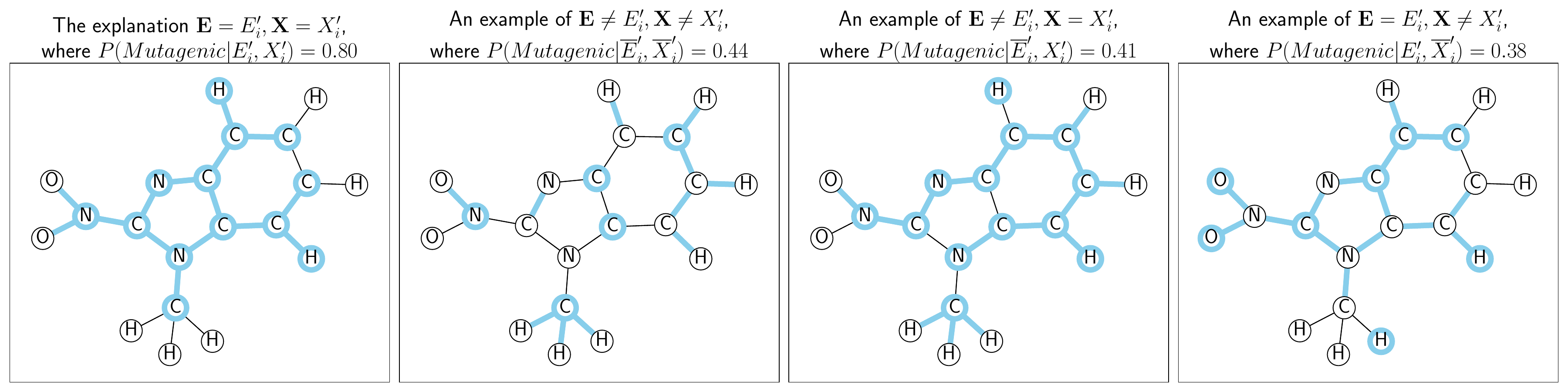}
    % \vspace{-1mm}
    \caption{Explanation of NSEG(PNS$^{e,f}$) on a Mutagenicity instance, where the explanation is highlighted in \textcolor{SkyBlue}{blue}.}
    \label{fig:qualitative_cf}
\end{figure*}

\subsection{Qualitative Studies}

Besides quantitative analysis, visual explanations can better help humans understand the decision-making process of GNNs. Hence, we present a series of qualitative studies as follows.

\mypara{Explanations of NSEG and Baselines} Figure~\ref{fig:qualitative_baselines} illustrates explanations obtained by different approaches in Mutagenicity and MSRC\_21 datasets. In Figure~\ref{fig:qualitative_baselines}~(a), we observe that NSEG(PNS) successfully identifies the well-known explanation, the nitro-group (-NO2), in the Mutagenicity dataset, while maintaining the integrity of the molecule by including the benzene ring and excluding the hydrogen bond (-H). Additionally, NSEG(PNS$^{e,f}$) generates a finer-grained explanation by identifying the node features of nitrogen (N), oxygen (O), and carbon (C) atoms. In Figure~\ref{fig:qualitative_baselines}~(b), the prediction of the instance graph is a scene about ``chair''. 
\chen{For edge-level explanations}, we observe that our approach NSEG(PNS) successfully identifies all edges among the node ``chair'' as the explanation, while GNNExplainer generates the explanation including not only edges among nodes ``chair'' but also edges among nodes ``void'', and CF-GNNExplainer only identifies the edges between nodes ``chair'' and nodes ``void''. \chen{For the node-level explanation generated by NSEG(PNS$^{e,f}$) (highlighted nodes in the rightmost column of Figure~\ref{fig:qualitative_baselines}~(b)), the inclusion of nodes related to the chair ensures the explanation’s sufficiency. However,  some ``void'' nodes may also increase the probability of predicting ``chair'' because they are strongly associated with the ``chair'' nodes in the training set. These scattered ``void'' nodes are unlikely to be shared by other ``chair''-predicted samples, making the explanation sufficient but unnecessary. This is expected, since the strength of our NSEG lies in its ability to generate explanations that are both  as sufficient and necessary as possible.}
% \yxzhu{However, we also observe our NSEG(PNS$^{e,f}$) identifies some redundant ``void'' node feature as the explanation, and the reason can be the sufficient aspect dominates the explanation searching process to some extent that leads to a sub-optimal solution.}

\mypara{Explanations of NSEG Variants} Figure~\ref{fig:qualitative_variants} illustrates the explanations obtained by the variants of NSEG, which are NSEG(PN$^{e}$), NSEG(PS$^{e}$), and NSEG(PNS$^{e}$) respectively.
Compared to the baselines, NSEG(PNS$^{e}$) identifies the well-known explanation, the amino-group (-NH2) \cite{c:gnnexplainer,c:pgexplainer}, while keeping the integrity of the molecule by including the benzene ring and excluding the hydrogen bond (-H). In contrast, the ablation NSEG(PN$^{e}$) only identifies the benzene ring (which is necessary but not sufficient), while NSEG(PS$^{e}$) identifies almost the whole graph (which is sufficient but not necessary).

\mypara{Explanations for Incorrect and Correct Predictions} 
To gain insights into the model's decision-making process, it is crucial to explain both correct and incorrect predictions.
Hence, we showcase two explanations of the instances with correct and incorrect predictions.  
The first instance in Figure~\ref{subfig:case_a} is a scene about ``water, boat...'', but the GNN incorrectly predicts ``road, building...''. The generated explanation suggests the reason behind the misclassification---the GNN paid more attention to the nodes ``road'', ``building'', and the edges among them.  The second instance in Figure~\ref{subfig:case_b} is a scene about ``dog'', which is correctly predicted. The generated explanation shows that more attention was paid to the nodes ``dog'' and edges among them, leading to the correct prediction.
Additionally, more qualitative studies are shown in \ref{sec:app_more_cases}.

\mypara{Explanations with both Necessity and Sufficiency} % Figure~\ref{fig:qualitative_cf} illustrates the necessary and sufficient explanation in an intuitive way. 
In Figure~\ref{fig:qualitative_cf} a compound is predicted as a ``mutagenic'', whose necessary and sufficient explanation $\mathbf{E}=E_i',\mathbf{X}=X_i'$ can produce the prediction ``mutagen'' with a high probability 0.93, showing the sufficiency of the explanation. Also, in the last three columns of Figure~\ref{fig:qualitative_cf}, we present one example for each of the cases: $(\mathbf{E}\neq E_i',\mathbf{X}\neq X_i')$, $(\mathbf{E}\neq E_i',\mathbf{X}=X_i')$, and $(\mathbf{E}=E_i',\mathbf{X}\neq X_i')$, which degrades the probabilities of producing ``mutagen'' to 0.44, 0.41, and 0.38, respectively. Such changes imply the necessity of the explanation.

\subsection{Sensitivity Analysis }
\begin{figure}[t!]
    \centering
    \includegraphics[scale=0.45]{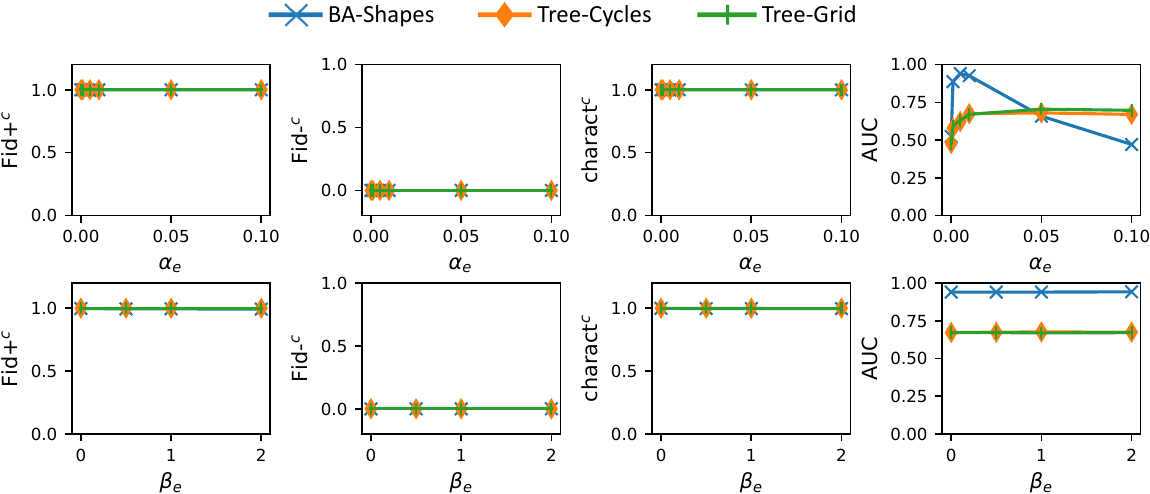}    
    \caption{\emph{Fid+}$^c$, \emph{Fid-}$^c$, \emph{charact}$^c$ and \emph{AUC} w.r.t. $\alpha_e$ on NSEG(PNS$^e$) across BA-Shapes, Tree-Cycles, and Tree-Grid datasets.}
    \label{fig:sensitivity_alpha}
\end{figure}

\begin{figure}[h]
	\centering
	\includegraphics[width=\columnwidth]{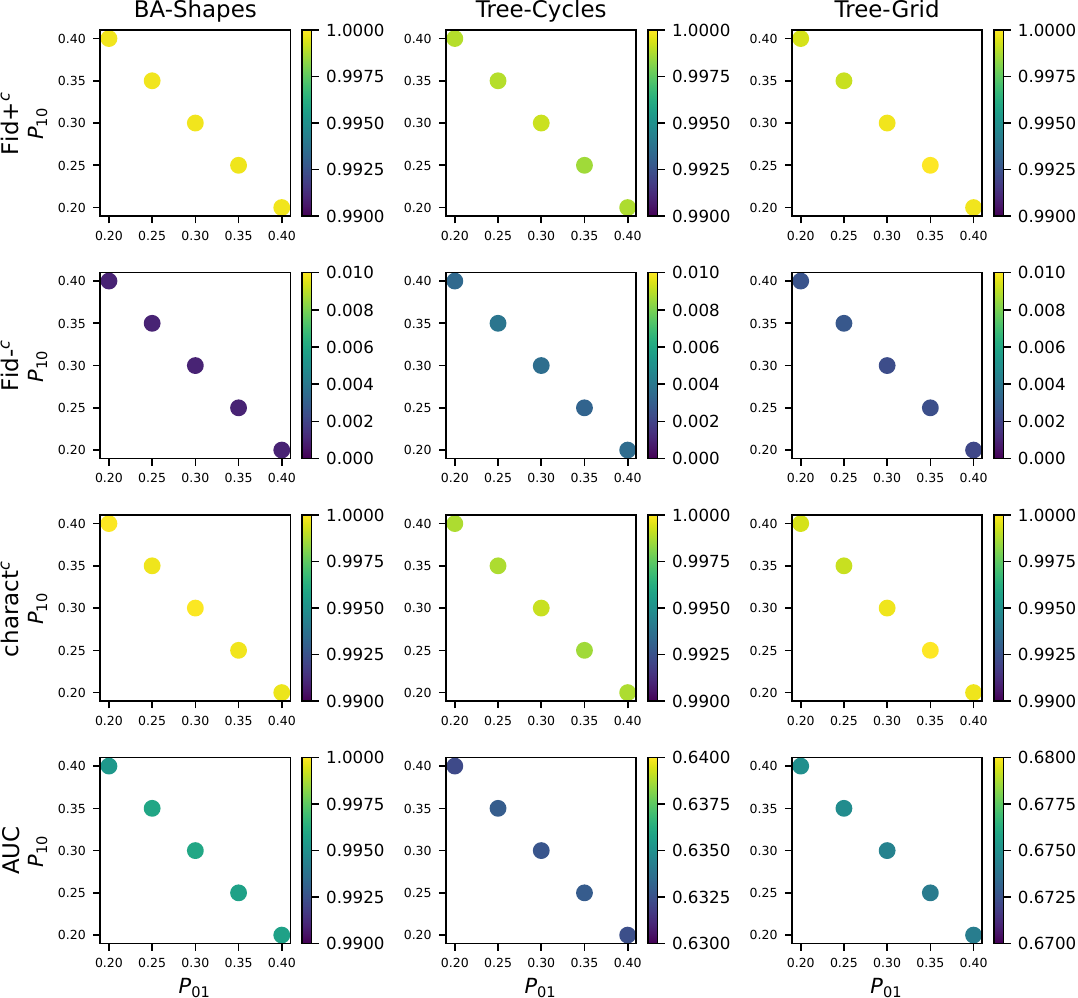} % 1.00208
  % \vspace*{-2.5mm}
	\caption{\emph{Fid+}$^c$, \emph{Fid-}$^c$, \emph{charact}$^c$ and \emph{AUC} w.r.t. $P_{01}$ and $P_{10}$ on NSEG(PNS$^{e,f}$) across BA-Shapes, Tree-Cycles, and Tree-Grid datasets, where $P_{00}=1-P_{01}-P_{10}$.}
 \label{fig:ablation_prior}
\end{figure}

\mypara{Sensitivity of $\alpha$ and $\beta$}
The hyper-parameter $\alpha$ and $\beta$ in Eq.~\eqref{eq:overall_loss} modulates the \emph{mask size} and \emph{mask entropy} regularization terms, the components also utilized in prior works \cite{c:gnnexplainer, c:pgexplainer, c:cf2, c:cfgnnexplainer}. Intuitively, the \emph{mask size} regularization operates as an $L_1$ penalization, compelling the explanation to emphasize the most salient portions of the input, while the \emph{mask entropy} forces the mask concentrating around a few
scalars. Within our experimental setup, we scrutinized the impact of $\alpha_e$ and $\beta_e$ on NSEG(PNS$^e$), where $\alpha_e$ and $\beta_e$ govern the \emph{mask size} and \emph{mask entropy} regularization of the edge mask $M_e$, respectively. The implications of varying $\alpha_e$ on NSEG(PNS$^e$) across the metrics of \emph{charact} and \emph{Recall@K} are depicted in Figure~\ref{fig:sensitivity_alpha}. Two key observations emerge from our analysis:
\begin{itemize}
    \item Regarding $\alpha_e$, metrics \emph{Fid+}$^c$, \emph{Fid-}$^c$ and \emph{charact}$^c$ remains largely impervious to alterations in $\alpha_e$, while in contrast, metric \emph{AUC} displays a pronounced sensitivity to $\alpha_e$.
    \item Regarding $\beta_e$, metrics $\alpha_e$, metrics \emph{Fid+}$^c$, \emph{Fid-}$^c$, \emph{charact}$^c$, and \emph{AUC} show non-sensitive to $\alpha_e$, evidenced by achieving stable results on these metrics.
\end{itemize}

\mypara{Sensitivity of the Priors}
\chen{We present the sensitivity analysis for the prior probabilities $P_{00}$, $P_{01}$, and $P_{10}$ (Eq.~\eqref{eq:p000110}) in our NSEG. Specifically, 
we first determined the appropriate ranges for \(P_{00}\), \(P_{01}\), and \(P_{10}\) around 1/3, based on the prior knowledge discussed in Section 4.3.2, using the range [0.2, 0.4]. We then fixed $P_{00}$ at 0.4 and linearly sampled 5 values for $P_{01}$ and $P_{10}$ at equal intervals within their respective ranges. As shown in Figure~\ref{fig:ablation_prior}, NSEG’s performance, evaluated using \emph{Fid+}\(^{c}\), \emph{Fid-}\(^{c}\), \emph{charact}\(^{c}\), and AUC, remains stable. This indicates that as long as the priors are within an appropriate range, NSEG’s performance is less sensitive to their values.}

\section{Conclusion and Discussion}
In this paper, we proposed NSEG, a GNN explanation model that is able to provide necessary and sufficient explanations for GNNs. Different from the existing approaches that generate either necessary or sufficient explanations, or a heuristic trade-off explanation between the two aspects, our objective is to maximize PNS that ensures the explanation is both necessary and sufficient. 
To overcome the intractability of identifying PNS, we derived a lower bound of PNS, and further propose a generalized SCM of GNN for estimating this lower bound. For optimization, we leverage continuous masks with sampling strategy to optimize the lower bound. Our empirical study demonstrates that NSEG can provide the most necessary and sufficient explanations and achieves state-of-the-art performance on various datasets.

\chen{However, the objective function of NSEG depends on the selection of priors, even though an appropriate range can be ensured based on prior knowledge.
In future work, we believe a more prior-free algorithm deserves further attention  to address the limitations of NSEG.
Despite these limitations, we believe our NSEG contributes to machine learning explainability, particularly in terms of necessity and sufficiency, and inspires the design of more interpretable models.
}

\bibliographystyle{elsarticle-harv}
\bibliography{main}

\newpage
\appendix

\section{Proof}

\subsection{Proof of Lemma~\ref{lemma:pns_lb}}
\begin{proof}
According to the definition of PNS in Definition~\ref{def:pns}, we have
\begin{equation*}
\small
\begin{aligned}
&\text{PNS}(\xi) = P(\mathbf{Y}_{\xi^{c}}\neq\hat{y},\mathbf{Y}_{\xi}=\hat{y})\\
=&P(\mathbf{Y}_{\xi^{c}}\neq\hat{y}) + P(\mathbf{Y}_{\xi}=\hat{y})-P(\mathbf{Y}_{\xi^{c}}\neq\hat{y}\vee\mathbf{Y}_{\xi}=\hat{y}),
\end{aligned}
\end{equation*}
since $P(\mathbf{Y}_{\xi^{c}}\neq\hat{y}\vee\mathbf{Y}_{\xi}=\hat{y})\leq 1$ and $P(\mathbf{Y}_{\xi^{c}}\neq\hat{y},\mathbf{Y}_{\xi}=\hat{y})\geq 0$, we have:   
\begin{equation*}
\small
\begin{aligned}
&\text{PNS}(\xi)\geq \mathop{\max}\{0,P(\mathbf{Y}_{\xi^{c}}\neq\hat{y})+P(\mathbf{Y}_{\xi}=\hat{y})-1\}.
\end{aligned}
\end{equation*}
\end{proof}

\subsection{Proof of Lemma~\ref{lemma:pns_tight}}
\label{sec:proof_lemma_2}

\begin{proof}
Since $(\mathbf{Y}_{\xi^{c}}=\hat{y})\vee(\mathbf{Y}_{\xi^{c}}\neq\hat{y})=true$, we have:
\begin{equation*}
\small
\begin{aligned}
&(\mathbf{Y}_{\xi}=\hat{y}) \\
=&(\mathbf{Y}_{\xi}=\hat{y})\wedge((\mathbf{Y}_{\xi^{c}}=\hat{y})\vee(\mathbf{Y}_{\xi^{c}}\neq\hat{y})) \\
=&((\mathbf{Y}_{\xi}=\hat{y})\wedge(\mathbf{Y}_{\xi^{c}}=\hat{y}))\vee((\mathbf{Y}_{\xi}=\hat{y})\wedge(\mathbf{Y}_{\xi^{c}}\neq\hat{y})),
\end{aligned}
\end{equation*}
and since $(\mathbf{Y}_{\xi}=\hat{y})\vee(\mathbf{Y}_{\xi}\neq\hat{y})=true$, when monotonicity holds, i.e., $(\mathbf{Y}_{\xi}\neq\hat{y})\wedge(\mathbf{Y}_{\xi^{c}}=\hat{y})=false$, we have:
\begin{equation*}
\small
\begin{aligned}
&(\mathbf{Y}_{\xi^{c}}=\hat{y})\\
=&(\mathbf{Y}_{\xi^{c}}=\hat{y})\wedge((\mathbf{Y}_{\xi}=\hat{y})\vee(\mathbf{Y}_{\xi}\neq\hat{y})) \\
=&((\mathbf{Y}_{\xi^{c}}=\hat{y})\wedge(\mathbf{Y}_{\xi}=\hat{y}))\vee((\mathbf{Y}_{\xi^{c}}=\hat{y})\wedge(\mathbf{Y}_{\xi}\neq\hat{y})) \\
=&((\mathbf{Y}_{\xi^{c}}=\hat{y})\wedge(\mathbf{Y}_{\xi}\neq\hat{y})),
\end{aligned}
\end{equation*}
then, combining those 2 equations, we have:
\begin{equation*}
\small
\begin{aligned}
&(\mathbf{Y}_{\xi}=\hat{y})\\
=&(\mathbf{Y}_{\xi^{c}}=\hat{y})\vee((\mathbf{Y}_{\xi}=\hat{y})\wedge(\mathbf{Y}_{\xi^{c}}\neq\hat{y}))
\end{aligned}
\end{equation*}
then take the probability form and use the disjointness of $\mathbf{Y}_{\xi^{c}}=\hat{y}$ and $\mathbf{Y}_{\xi^{c}}\neq\hat{y}$:
\begin{equation*}P(\mathbf{Y}_{\xi}=\hat{y})=P(\mathbf{Y}_{\xi^{c}}=\hat{y})+P(\mathbf{Y}_{\xi}=\hat{y},\mathbf{Y}_{\xi^{c}}\neq\hat{y})
\end{equation*}
then:
\begin{equation*}
\small
\begin{aligned}
\text{PNS}(\xi)=&P(\mathbf{Y}_{\xi}=\hat{y})-P(\mathbf{Y}_{\xi^{c}}=\hat{y}) \\
=&P(\mathbf{Y}_{\xi}=\hat{y})+P(\mathbf{Y}_{\xi^{c}}\neq\hat{y})-1
\end{aligned}
\end{equation*}
\end{proof}

\subsection{Proof of Proposition~\ref{prop:scm_gnn}}
\label{sec:proof_scm_gnn}
\begin{proof}
In a graph neural network, each node's hidden representation in each hidden layer can be written as functions of entries of graph and all nodes' representations in the previous layer, i.e., 
\begin{equation*}
\small
\forall{v}\forall{k}(\mathbf{h}_{i}^{(k)}=f_{\theta}^{(k)}(\mathbf{E}_{*,i},\mathbf{H}^{(k-1)})),
\end{equation*}
where $\mathbf{h}_{v}^{(k)}$ denotes the hidden representation of node $v$ in $k$-th layer, $\mathbf{H}^{(k-1)})$ denotes the hidden representations of all nodes in $(k-1)$-th layer, $\mathbf{E}_{*,i}$ denotes the entries from all nodes to node $i$. Then for all nodes' representations, we have,
\begin{equation*}
\small
\forall{k}(\mathbf{H}^{(k)}=f_{\theta}^{(k)}(\mathbf{E},\mathbf{H}^{(k-1)})),
\end{equation*}
where $\mathbf{E}$ denotes the edges of the graph. 

The output variable $\mathbf{Y}$ is determined by all nodes' representations in $l$-th layer by a read-out function $f_{\theta}^{(l+1)}$ (specifically, a node-level read-out for node classification task and a graph-level read-out for graph classification task), i.e., 
\begin{equation*}
\small
\mathbf{Y}=f_{\theta}^{(l+1)}(\mathbf{H}^{(l)}).
\end{equation*}

And for the input variables $\mathbf{E}$ and $\mathbf{X}$, they are determined by a set of exogenous $\mathbf{U}$, i.e., 
\begin{equation*}
\small
\mathbf{E},\mathbf{X}=f^{(0)}(\mathbf{U}).
\end{equation*}

Thus, the structure of those variables can be equivalently expressed by a SCM $M(\{\mathbf{X},\mathbf{E},\mathbf{H}^{(1)},...,\mathbf{H}^{(l)},\mathbf{Y}\},\mathbf{U},\{f^{(0)},f_{\theta}^{(1)},...,f_{\theta}^{(l+1)}\})$.
\end{proof}

\subsection{Proof of Proposition~\ref{prop:scm_gnn_reduced}}
\label{sec:proof_scm_gnn_reduced}
\begin{proof}
Start from the output variable $\mathbf{Y}$, the corresponding causal mechanism function $\mathbf{Y}=f_{\theta}^{(l+1)}(\mathbf{H}^{(l)})$ can be substituted as
\begin{equation*}
\small
\mathbf{Y}=f_{\theta}^{(l+1)}(f_{\theta}^{(l)}(\mathbf{E},f_{\theta}^{(l-1)}(\mathbf{E}, ...f_{\theta}^{(1)}(\mathbf{E},\mathbf{X})))).
\end{equation*}

Thus, we can obtain a new causal mechanism $f_{\theta}$ that is a mapping from $\mathbf{E},\mathbf{X}$ to $\mathbf{Y}$, i.e.,
\begin{equation*}
\small
\mathbf{Y}=f_{\theta}(\mathbf{E},\mathbf{X}).
\end{equation*}

Hence, we can obtain a reduced SCM $M(\{\mathbf{E},\mathbf{X},\mathbf{Y}\},\mathbf{U},\{f^{(0)},f_{\theta}\})$.
\end{proof}

\subsection{Proof of Proposition~\ref{prop:pn_ps_reduced}}
\label{sec:app_proof_lb_pnps}
\begin{proof}
First, we can derive lower bounds of both PN and PS as follows.

\begin{equation*}
\small
\begin{aligned}
\text{PN}(\xi)= &P(\mathbf{Y}_{\xi^{c}}\neq\hat{y}|\xi,\mathbf{Y}=\hat{y}) \\
=&\frac{P(\mathbf{Y}_{\xi^{c}}\neq\hat{y},\xi,\mathbf{Y}=\hat{y})}{P(\xi,\mathbf{Y}=\hat{y})}\\
\overset{Propostion~2}{=}&\frac{P(\mathbf{Y}_{\xi^{c}}\neq\hat{y},\xi,\mathbf{Y}_{\xi}=\hat{y})}{P(\xi,\mathbf{Y}=\hat{y})}\\
=&\frac{P(\mathbf{Y}_{\xi^{c}}\neq\hat{y},\mathbf{Y}_{\xi}=\hat{y})P(\xi)}{P(\xi,\mathbf{Y}=\hat{y})}\\
=&\frac{P(\mathbf{Y}_{\xi^{c}}\neq\hat{y},\mathbf{Y}_{\xi}=\hat{y})}{P(\mathbf{Y}=\hat{y}|\xi)}\\
\geq&\frac{\mathop{\max}\{0,P(\mathbf{Y}_{\xi^{c}}\neq\hat{y})+P(\mathbf{Y}_{\xi}=\hat{y})-1\}}{P(\mathbf{Y}=\hat{y}|\xi)}\\
=&\mathop{\max}\{0,\frac{P(\mathbf{Y}_{\xi^{c}}\neq\hat{y})-1}{P(\mathbf{Y}=\hat{y}|\xi)}+1\}
\end{aligned}
\end{equation*}

\begin{equation*}
\small
\begin{aligned}
\text{PS}(\xi)=&P(\mathbf{Y}_{\xi}=\hat{y}|\xi^{c},\mathbf{Y}\neq\hat{y}) \\
=&\frac{P(\mathbf{Y}_{\xi}=\hat{y},\xi^{c},\mathbf{Y}\neq\hat{y})}{P(\xi^{c},\mathbf{Y}\neq\hat{y})}\\
\overset{Propostion~2}{=}&\frac{P(\mathbf{Y}_{\xi}=\hat{y},\xi^{c},\mathbf{Y}_{\xi^{c}}\neq\hat{y})}{P(\xi^{c},\mathbf{Y}\neq\hat{y})}\\
=&\frac{P(\mathbf{Y}_{\xi}=\hat{y},\mathbf{Y}_{\xi^{c}}\neq\hat{y})P(\xi^{c})}{P(\xi^{c},\mathbf{Y}\neq\hat{y})}\\
=&\frac{P(\mathbf{Y}_{\xi}=\hat{y},\mathbf{Y}_{\xi^{c}}\neq\hat{y})}{P(\mathbf{Y}\neq\hat{y}|\xi^{c})}\\
\geq&\frac{\mathop{\max}\{0,P(\mathbf{Y}_{\xi^{c}}\neq\hat{y})+P(\mathbf{Y}_{\xi}=\hat{y})-1\}}{P(\mathbf{Y}\neq\hat{y}|\xi^{c})}\\
=&\mathop{\max}\{0,\frac{P(\mathbf{Y}_{\xi}=\hat{y})-1}{P(\mathbf{Y}_{\xi^{c}}\neq\hat{y})}+1\}
\end{aligned}
\end{equation*}

Altering the event $\xi$ with the explanation event $(\mathbf{E}=E_i',\mathbf{X}=X_i')$, and under the SCM depict in Proposition~\ref{prop:scm_gnn_reduced}, the lower bounds of PN and PS after intervening on both $\mathbf{E}$ and $\mathbf{X}$ (replacing causal mechanism from $\mathbf{U}$ to $\mathbf{E}$ and $\mathbf{X}$), we have:
\begin{equation*}
\small
\begin{aligned}
\text{PN}_{lb}^{e,f}(E_i',X_i')&=\mathop{\max}\{0,\frac{P_{\theta}(\mathbf{Y}\neq\hat{y}|(\mathbf{E}=E_i',\mathbf{X}=X_i')^{c})-1}{P_{\theta}(\mathbf{Y}=\hat{y}|\mathbf{E}=E_i',\mathbf{X}=X_i')}\!+\!1\},\\
\text{PS}_{lb}^{e,f}(E_i',X_i')&=\mathop{\max}\{0,\frac{P_{\theta}(\mathbf{Y}=\hat{y}|\mathbf{E}=E_i',\mathbf{X}=X_i')-1}{P_{\theta}(\mathbf{Y}\neq\hat{y}|(\mathbf{E}=E_i',\mathbf{X}=X_i')^{c})}\!+\!1\}.
\end{aligned}
\end{equation*}

\end{proof}

\section{Optimization Details of the PN and PS Lower Bounds}\label{sec:detail_lbd}
\chen{In practice, directly optimizing the lower bounds of PN and PS in Proposition~\ref{prop:pn_ps_reduced} for NSEG(PN$^e$) and NSEG(PS$^e$) often leads to suboptimal solutions, manifested by the dominance of the denominator in the optimization, while the numerator remains nearly unchanged. To alleviate this issue, since these two bounds are not the primary focus of this work, we simply fix their denominators and optimize only the numerators to avoid suboptimal solutions. Mathematically, combining the previously discussed continuous mask and sampling strategy, the objective functions for optimizing $\text{PN}_{lb}^{e,f}$ and $\text{PS}_{lb}^{e,f}$ are as follows.}
\begin{equation*}
\small
\begin{aligned}
\label{eq:pn_lb_ef_op_final}
\text{PN}_{lb}^{e,f}=&-P_{00}\mathbb{E}_{\epsilon_e,\epsilon_f}[f_{\theta}^{\hat{y}}((1\!-\!M_{e}+\epsilon_{e})\!\odot\! E_i,(1-M_{f}+\epsilon_{f})\!\odot\! X_i)]\\
&-P_{01}\mathbb{E}_{\epsilon_e}[f_{\theta}^{\hat{y}}((1-M_{e}+\epsilon_{e})\odot E_i,M_{f}\odot X_i)]\\
&-P_{10}\mathbb{E}_{\epsilon_f}[f_{\theta}^{\hat{y}}(M_{e}\odot E_i,(1-M_{f}+\epsilon_{f})\odot X_i)]+1,
\end{aligned}
\end{equation*}
\begin{equation*}
\label{eq:ps_lb_ef_op_final}
\small
\text{PS}_{lb}^{e,f}=f_{\theta}^{\hat{y}}(M_{e}\odot E_i,M_{f}\odot X_i).
\end{equation*}
\chen{For tasks requiring higher precision in the PN and PS lower bounds, we can apply existing approaches such as alternating optimization of the numerator and denominator~\cite{boyd2011distributed}.}

\section{Details of Experiment Setup}
\label{sec:app_detail}

\subsection{GNN Training Setup}
\label{sec:gnn_training}
In this section, we introduce the training setup and training result of the GNN we explain.
For the GNN training, we employ three layers of Graph Convolutional Networks (GCNs) \cite{c:gcn} with ReLU activation. 
For node classification task, the dimensions of the hidden layers are 16, 32, and 16 respectively, followed by a fully connected layer as the output layer. For graph classification task, the dimensions of the hidden layer in graph classification task are 16, 32, and 16, and after the last GCN layer, a sum-based read-out is used to obtain a graph representation, followed by a fully connected layer as the output layer. 
For all datasets, we split train/val/test with 80~\%/10~\%/10~\%. 
The hyperparameter setting and test accuracy of the trained GNN model are shown in Table~\ref{tab:training_res}.

\begin{table*}[t]
    \centering
    \caption{The hyperparameter setting and test accuracy of GNN.}
    \scalebox{0.8}{
\begin{tabular}{lcccccc}
\hline
                 & BA-Shapes & Tree-Cycles & Tree-Grid & BA2Motif & Mutagenicity & MSRC\_21 \\ \hline
lr               & 0.001     & 0.001       & 0.001     & 0.01     & 0.001        & 0.001    \\
\# of GNN layers & 3         & 3           & 3         & 3        & 3            & 3        \\
\# of epochs     & 2000      & 2000        & 2000      & 2000     & 500          & 500      \\
dropout          & 0         & 0           & 0         & 0        & 0.5          & 0.5      \\
optimizer        & Adam      & Adam        & Adam      & Adam     & Adam         & Adam     \\
weight decay     & 0         & 0           & 0         & 0        & 5e-4         & 5e-4     \\ \hline
accuracy         & 0.957     & 0.903       & 0.927     & 1.0      & 0.761        & 0.983    \\ \hline
\end{tabular}
    }
    \label{tab:training_res}
\end{table*}

\subsection{Baselines}
\label{sec:baselines}
\mypara{GNNExplainer}
The goal of GNNExplainer is to identify a subgraph $G_i'$ and the associated features $X_i'$ that are important to GNN's prediction $\hat{y}$. The objective of GNNExplainer is to maximize the mutual information between the subgraph explanation and the GNN model outcome. Intuitively, if knowing the information of the subgraph $G_i'$ and its associated features $X_i'$ can reduce the uncertainty of $\mathbf{Y}$, then $G_i'$ and $X_i'$ are good explanations for the GNN prediction. Equivalently, the mutual information objective captures the sufficiency aspect in producing an explanation. 
\begin{equation*}
\small
\mathop{\max}\limits_{G_i',X_i'}{MI(\mathbf{Y};(G_i',X_i'))} 
\end{equation*}
where $MI$ quantifies the change in the probability of the prediction $\hat{y}$ when the input is limited to $(G_i',X_i')$. In particular, GNNExplainer converts the discrete optimization problem into continuous optimization and leverages trainable edge mask $M_e$ and feature mask $M_f$ in optimization, where the objective is derived as follow.
\begin{equation*}
\small
\mathop{\min}\limits_{M_{e},M_{f}}{H(\mathbf{Y}|\mathbf{G}=G_i\odot M_{e},\mathbf{X}=X_i\odot M_{f})}
\end{equation*}

Notably, the $X_i'$ in GNNExplainer is quite different from the $X_i'$ in NSEG, where the former is a feature dimension-wise explanation while the latter is a node-wise explanation.
In implementations, we adopt the original settings where for node classification task the $l$-hop subgraph of the target node is extracted, where $l$ denotes the number of graph convolution layers. For hyper-parameter that controls the \emph{mask size} regularization, we select [0.01, 0.01, 0.01, 0.005, 0.005] for BA-Shapes, Tree-Cycles, Tree-Grid, Mutagenicity, and MSRC\_21, respectively. For hyper-parameter that controls the \emph{entropy} regularization, we set it to 1.0 for all datasets.

\mypara{PGExplainer}
Though PGExplainer shares the same objective with GNNExplainer, i.e., maximizing the mutual information between the subgraph explanation and the GNN outcome, PGExplainer spends way less time in inference than GNNExplainer. To achieve this, the key idea of PGExplainer is to learn a mapping from the graph representation space to subgraph space parameterized by $\omega$, and the subgraphs are sampled from the distribution $q(\omega)$. The parameterized mapping gives PGExplainer the inductive ability, which means the explainer once trained, it does not need to be retrained to explain different instances. In particular, the objective of PGExplainer is shown as follows.
\begin{equation*}
\small
\mathop{\min}\limits_{\omega}{\mathbb{E}_{G_i'\sim{q(\omega)}}H(\mathbf{Y}|G_i')},
\end{equation*}
where the sampling process $G_i'\sim{q(\omega)}$ is approximated via determinant function of parameters $\omega$, temperature $\tau$, and an independent random variable $\epsilon\sim U(0,1)$:
\begin{equation*}
\small
G_i'=f_{\omega}(G_i, \tau, \epsilon).
\end{equation*}

In implementations, we adopt the original annealing schedule of $\tau$, i.e., $\tau_t=(\tau_T/\tau_0)^t$, where $\tau_0$ and $\tau_T$ are set to 5.0 and 2.0, respectively. For node classification task the $l$-hop subgraph of the target node is extracted, where $l$ denotes the number of graph convolution layers. For the hyper-parameter that controls the \emph{mask size} regularization, we select [0.01, 0.01, 0.01, 0.05, 0.05] for BA-Shapes, Tree-Cycles, Tree-Grid, Mutagenicity, and MSRC\_21, respectively.
For the hyper-parameter that controls the \emph{entropy} regularization, we set it to 1.0 for all datasets.

\mypara{PGM-Explainer}
A Probabilistic Graphical Model (PGM) is utilized as a surrogate in PGM-Explainer to generate explanations for GNN. 
The goal of PGM-Explainer is to learn a Bayesian network upon the local perturbation-prediction dataset to identify significant nodes.
Specifically, to obtain the local perturbation-prediction dataset, PGM-Explainer randomly perturbs the node features of several random nodes, then records a random variable indicating whether its features are perturbed and its influence on the GNN predictions.
After obtaining the local dataset, the Grow-Shrink (GS) algorithm is leveraged to reduce the size of the local dataset, then an interpretable Bayesian network is employed to fit the local dataset and to explain the GNN model.
For hyper-parameters settings, we set the probability of perturbation to 0.5, the prediction difference threshold to 0.1, the number of samples to 50, and the p-value for the conditional independence test to 0.05.

\mypara{CF-GNNExplainer}
Different from GNNExplainer and PGExplainer whose objective is to generate subgraphs that are especially relevant for a particular prediction, the goal of CF-GNNExplainer is to generate a counterfactual explanation that can flip the model prediction. The nature of counterfactual explanation \cite{c:cf_gnnexplainer} is that the explanation can flip the model prediction subject to a minimal perturbation on data, which is equivalent to the nature of the necessary explanation. To generate the counterfactual explanation $G_i'$, the objective of CF-GNNExplainer is given as follows.
\begin{equation*}
\small
\mathop{\max}\limits_{G_i'}{\mathcal{L}_{pred}(G_i,G_i')+\beta\mathcal{L}_{dist}(G_i,G_i')},
\end{equation*}
where $\mathcal{L}_{pred}(G_i,G_i')$ stands for the prediction loss aiming to obtain $G_i'$ that can flip the model prediction, and $\mathcal{L}_{dist}(G_i,G_i')$ stands for the distance loss aiming to achieve minimum perturbation.
In implementations, we adopt the original settings where for node classification task the $(l+1)$-hop subgraph of the target node is extracted, where $l$ denotes the number of graph convolution layers.
For $\beta$ we select [0.05, 0.05, 0.05, 0.01, 0.01] for BA-Shapes, Tree-Cycles, Tree-Grid, Mutagenicity, and MSRC\_21, respectively.

\mypara{CF$^2$}
CF$^2$ produces a trade-off explanation between necessity and sufficiency by taking insights into counterfactual and factual reasoning from causal inference theory. The objective of CF$^2$ is to minimize the complexity of the explanation, subject to the explanation strength being strong enough. The explanation strength can be divided into two parts, the counterfactual explanation strength $S_c$ and the factual explanation strength $S_f$, and a hyperparameter is introduced to control the trade-off between the two. Formally, the objective of CF$^2$ is shown as follows.
\begin{equation*}
\small
\mathop{\min}\limits_{G_i',X_i'}{C(G_i')}\quad s.t. \quad \alpha S_f(G_i',X_i') + (1-\alpha) S_c(G_i',X_i') > \lambda,
\end{equation*}
where $C(G_i',X_i')$ measures the complexity of the explanation $G_i'$, and $X_i'$ is similar to the $X_i'$ of GNNExplainer as we mentioned before. For implementations, we adopt the original settings where $\alpha$ is set to 0.6 for all datasets, and $\lambda$ is set to [500, 500, 500, 1000, 1000] for BA-Shapes, Tree-Cycles, Tree-Grid, Mutagenicity, and MSRC\_21, respectively.
Also for node classification task the $l$-hop subgraph of the target node is extracted, where $l$ denotes the number of graph convolution layers.

\subsection{Hardware and Software}
For hardware, our experiments are conducted on on a Linux machine with Nvidia GeForce RTX 2080 Ti with 11 GB memory.
For software, we implement our NSEG in Deep Graph Library (DGL) with Pytorch. For the implementations of the baseline approaches, we follow and modify the following codes: GNNExplainer\footnote{\url{https://github.com/RexYing/gnn-model-explainer}}, PGExplainer\footnote{\url{https://github.com/flyingdoog/PGExplainer}}, 
PGM-Explainer\footnote{\url{https://github.com/mims-harvard/GraphXAI/tree/main/graphxai/explainers/pgm_explainer}},
CF-GNNExplainer\footnote{\url{https://github.com/a-lucic/cf-gnnexplainer}}, and CF$^2~$\footnote{\url{https://github.com/chrisjtan/gnn_cff}}.

\section{More Experimental Results}
\subsection{Result on GIN Architecture}

In this subsection, we showcase the additional experimental results on GIN \cite{c:gin} architecture besides the GCN architects, where the results of \emph{Fid+$^c$}, \emph{Fid-$^c$} and \emph{charact$^c$} are given in Table~\ref{tab:fid_gin}, and the results of \emph{Recall@K} and \emph{ROC-AUC} are given in Table~\ref{tab:acc_gin}.
We observe that our NSEG achieves the best \emph{Fid+$^c$}, \emph{Fid-$^c$} and \emph{charact$^c$} compared with other approaches except for \emph{Fid-$^c$} and \emph{charact$^c$} on Tree-Grid dataset (NSEG also achieve the second best). Besides, our NSEG achieves the best \emph{Recall@K} on BA-Shapes and Tree-Grid, while achieving the second winner on the remaining metrics.

\begin{table*}[]
    \centering
    \caption{Comparison of \emph{Fid+$^c$}(\%), \emph{Fid-$^c$} (\%) and \emph{charact$^c$} (\%) of the explanations on GIN architecture. Mean and standard deviation are reported.}
    \addtolength{\tabcolsep}{-1.4mm}
\scalebox{0.8}{
\begin{tabular}{lccccccccc}
\hline
\multicolumn{10}{c}{Node Classification}                                                                                                                                                                                                                                                                             \\ \hline
\multicolumn{1}{l|}{}               & \multicolumn{3}{c|}{BA-Shapes}                                                                  & \multicolumn{3}{c|}{Tree-Cycles}                                                                & \multicolumn{3}{c}{Tree-Grid}                                              \\
\multicolumn{1}{l|}{}               & Fid+$^c$ ($\uparrow$) & Fid-$^c$ ($\downarrow$) & \multicolumn{1}{c|}{charact$^c$ ($\uparrow$)} & Fid+$^c$ ($\uparrow$) & Fid-$^c$ ($\downarrow$) & \multicolumn{1}{c|}{charact$^c$ ($\uparrow$)} & Fid+$^c$ ($\uparrow$) & Fid-$^c$ ($\downarrow$) & charact$^c$ ($\uparrow$) \\ \hline
\multicolumn{1}{l|}{Random}         & 62.50±0.00   & 62.45±0.10              & \multicolumn{1}{c|}{46.91±0.08}               & 45.56±2.76            & 47.44±3.17              & \multicolumn{1}{c|}{48.80±2.92}               & 98.75±0.00            & \textbf{98.75±0.00}     & \textbf{2.47±0.00}       \\
\multicolumn{1}{l|}{GuidedBP}       & 61.75±0.00            & 62.50±0.00              & \multicolumn{1}{c|}{46.66±0.00}               & 61.39±0.00            & 63.61±0.00              & \multicolumn{1}{c|}{45.69±0.00}               & 98.75±0.00            & \textbf{98.75±0.00}     & \textbf{2.47±0.00}       \\
\multicolumn{1}{l|}{GNNExplainer}   & 62.00±0.00            & 62.00±0.00              & \multicolumn{1}{c|}{47.12±0.00}               & 96.48±0.69            & 9.07±1.25               & \multicolumn{1}{c|}{93.62±0.98}               & 98.75±0.00            & \textbf{98.75±0.00}     & \textbf{2.47±0.00}       \\
\multicolumn{1}{l|}{PGExplainer}    & 48.50±10.49           & 62.50±0.00              & \multicolumn{1}{c|}{41.78±4.28}               & 44.07±26.47           & 42.96±29.19             & \multicolumn{1}{c|}{49.60±27.96}              & 95.51±2.67            & \textbf{98.75±0.00}     & \textbf{2.47±0.00}       \\
\multicolumn{1}{l|}{CFGNNExplainer} & \textbf{62.75±0.50}   & 62.50±0.00              & \multicolumn{1}{c|}{46.94±0.14}               & 98.33±0.00            & 81.81±0.14              & \multicolumn{1}{c|}{30.71±0.20}               & 98.47±0.28            & \textbf{98.75±0.00}     & \textbf{2.47±0.00}       \\
\multicolumn{1}{l|}{CF$^2$}         & 62.50±0.00            & \textbf{58.00±0.00}     & \multicolumn{1}{c|}{\textbf{50.24±0.00}}      & \textbf{100.00±0.00}  & \textbf{0.00±0.00}      & \multicolumn{1}{c|}{\textbf{100.00±0.00}}     & 98.75±0.00            & \textbf{98.75±0.00}     & \textbf{2.47±0.00}       \\
\multicolumn{1}{l|}{NSEG(PNS$^e$)}  & 62.50±0.00            & \textbf{58.00±0.00}     & \multicolumn{1}{c|}{\textbf{50.24±0.00}}      & \textbf{100.00±0.00}  & \textbf{0.00±0.00}      & \multicolumn{1}{c|}{\textbf{100.00±0.00}}     & \textbf{100.00±0.00}  & \textbf{98.75±0.00}     & \textbf{2.47±0.00}       \\ \hline
\multicolumn{10}{c}{Graph Classification}                                                                                                                                                                                                                                                                            \\ \hline
\multicolumn{1}{c|}{}               & \multicolumn{3}{c|}{BA2 Motif}                                                                  & \multicolumn{3}{c|}{Mutagenicity}                                                               & \multicolumn{3}{c}{MSRC\_21}                                               \\
\multicolumn{1}{l|}{}               & Fid+$^c$ ($\uparrow$) & Fid-$^c$ ($\downarrow$) & \multicolumn{1}{c|}{charact$^c$ ($\uparrow$)} & Fid+$^c$ ($\uparrow$) & Fid-$^c$ ($\downarrow$) & \multicolumn{1}{c|}{charact$^c$ ($\uparrow$)} & Fid+$^c$ ($\uparrow$) & Fid-$^c$ ($\downarrow$) & charact$^c$ ($\uparrow$) \\ \hline
\multicolumn{1}{l|}{Random}         & \textbf{49.50±0.00}   & 49.50±0.00              & \multicolumn{1}{c|}{49.99±0.00}               & 55.84±1.38            & 54.96±1.49              & \multicolumn{1}{c|}{49.85±1.20}               & 2.88±0.63             & 93.04±0.98              & 3.98±0.52                \\
\multicolumn{1}{l|}{GuidedBP}       & \textbf{49.50±0.00}   & 49.50±0.00              & \multicolumn{1}{c|}{49.99±0.00}               & 66.80±0.00            & 29.60±0.00              & \multicolumn{1}{c|}{68.55±0.00}               & 39.2±0.00             & 34.00±0.00              & 49.19±0.00               \\
\multicolumn{1}{l|}{GNNExplainer}   & 43.67±0.24            & 49.50±0.00              & \multicolumn{1}{c|}{46.84±0.14}               & 92.67±0.50            & 0.53±0.19               & \multicolumn{1}{c|}{95.95±0.27}               & 59.47±0.69            & 2.80±0.45               & 73.79±0.59               \\
\multicolumn{1}{l|}{PGExplainer}    & \textbf{49.50±0.00}   & 49.50±0.00              & \multicolumn{1}{c|}{49.99±0.00}               & 61.47±0.38            & 58.27±0.38              & \multicolumn{1}{c|}{49.71±0.35}               & 38.27±10.01           & 24.40±4.73              & 50.56±9.67               \\
\multicolumn{1}{l|}{CFGNNExplainer} & 48.00±1.08            & 49.50±0.00              & \multicolumn{1}{c|}{49.21±0.57}               & 66.40±1.99            & 14.53±0.19              & \multicolumn{1}{c|}{74.72±1.28}               & 15.73±0.69            & 97.20±0.00              & 4.75±0.03                \\
\multicolumn{1}{l|}{CF$^2$}         & \textbf{49.50±0.00}   & 49.50±0.00              & \multicolumn{1}{c|}{49.99±0.00}               & \textbf{100.00±0.00}  & \textbf{0.00±0.00}      & \multicolumn{1}{c|}{\textbf{100.00±0.00}}     & 87.60±0.00            & 1.20±0.00               & 92.86±0.00               \\
\multicolumn{1}{l|}{NSEG(PNS$^e$)}  & \textbf{49.50±0.00}   & \textbf{23.50±0.71}     & \multicolumn{1}{c|}{\textbf{60.11±0.22}}      & {96.13±0.19}  & {0.40±0.00}      & \multicolumn{1}{c|}{{97.84±0.10}}     & \textbf{90.40±0.00}            & \textbf{0.40±0.00}               & \textbf{94.78±0.00}               \\ \hline
\end{tabular}
}
    \label{tab:fid_gin}
\end{table*}

\begin{table*}[]
    \centering
    \caption{Comparison of \emph{Recall@K} (\%) and \emph{ROC-AUC} (\%) of the explanations on GIN architecture. Mean and standard deviation are reported.}
    \addtolength{\tabcolsep}{-1.4mm}

\resizebox{\linewidth}{!}{
\begin{tabular}{l|cc|cc|cc|cc|cc}
\hline
               & \multicolumn{2}{c|}{BA-Shapes} & \multicolumn{2}{c|}{Tree-Cycles} & \multicolumn{2}{c|}{Tree-Grid} & \multicolumn{2}{c|}{BA2Motif} & \multicolumn{2}{c}{Mutagenicity} \\
               & Recall@K       & ROC-AUC       & Recall@K        & ROC-AUC        & Recall@K       & ROC-AUC       & Recall@K       & ROC-AUC      & Recall@K        & ROC-AUC        \\ \hline
Random         & 33.33±0.38     & 50.09±0.29    & 66.94±0.31      & 50.82±0.38     & 64.04±0.26     & 49.82±0.21    & 22.22±1.23     & 50.46±1.00   & 40.20±0.87      & 49.71±0.85     \\
GuidedBP       & 77.77±0.00     & 98.98±0.00    & 76.85±0.00      & 73.04±0.00     & {74.83±0.00}     & {71.31±0.00}    & 36.90±0.00     & 73.28±0.00   & 76.67±0.00      & 80.07±0.00     \\
GNNExplainer   & 62.58±0.00     & 74.89±0.00    & 67.40±0.24      & 52.38±1.09     & 73.40±0.11     & 65.24±0.02    & 20.15±0.42     & 51.59±0.09   & 35.56±0.65      & 53.64±0.07     \\
PGExplainer    & {82.54±8.43}     & {99.14±0.47}    & {78.53±5.87}      & {75.68±5.39}     & 67.20±1.21     & 59.48±7.59    & 24.63±10.33    & 63.66±5.49   & 40.84±24.44     & 50.06±25.12    \\
CFGNNExplainer & 68.59±0.24     & 62.57±0.11    & 70.98±0.24      & 64.91±0.11     & 69.82±0.18     & 62.69±0.17    & 27.50±0.10     & 57.31±0.02   & 76.66±0.20      & 78.54±0.07     \\
CF$^2$         & 61.67±0.10     & 65.76±0.09    & 74.68±0.25      & 72.76±0.00     & 64.26±0.07     & 54.49±0.04    & 18.86±0.13     & 55.84±0.03   & 53.70±0.28      & 62.80±0.29     \\
NSEG(PNS$^e$)  & 65.43±0.05     & 79.63±0.03    & 78.35±0.07      & 73.31±0.05     & 65.23±0.06     & 58.71±0.04    & {37.35±0.73}     & 70.74±0.51   & {78.13±0.30}     & 74.82±0.19     \\ \hline
\end{tabular}
}
    \label{tab:acc_gin}
\end{table*}

\subsection{More Qualitative Studies}
\label{sec:app_more_cases}
\begin{figure*}[h]
    \centering
    \subfigure[]{
    \includegraphics[scale=0.11]{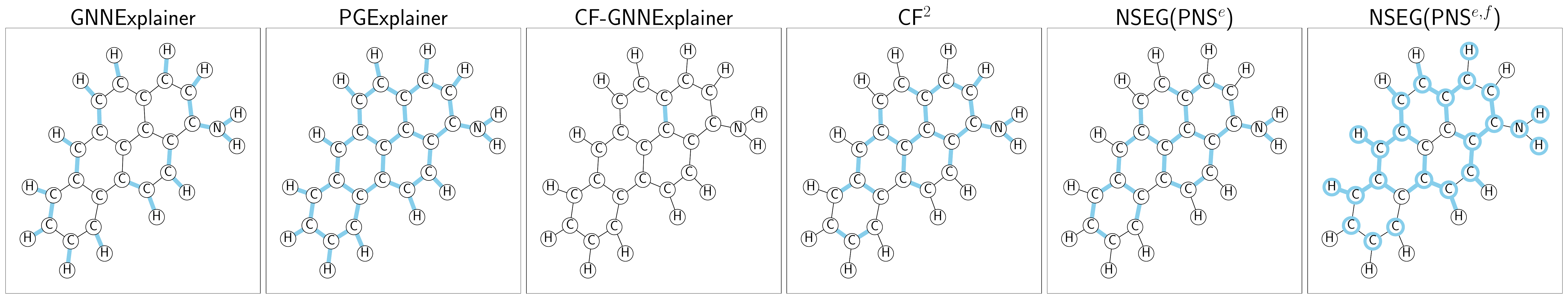}
    }
    \vspace{-1.8mm}
    \subfigure[]{
    \includegraphics[scale=0.11]{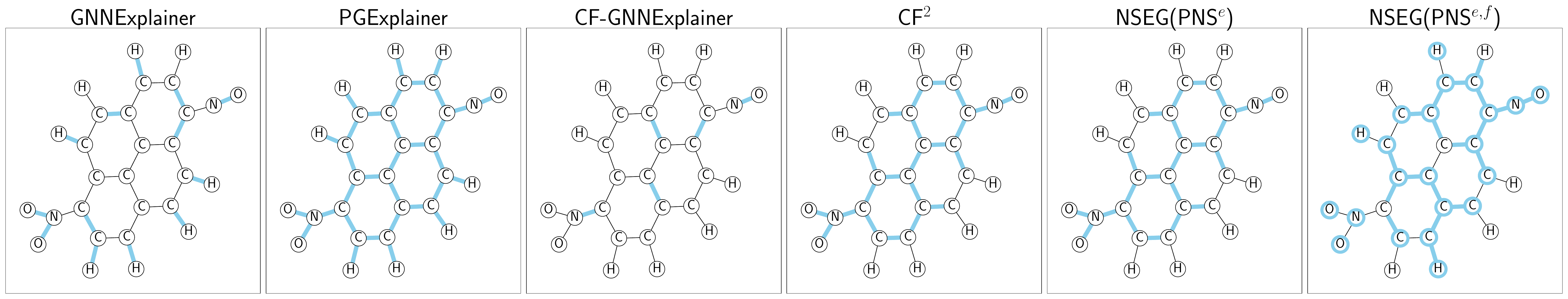}
    }
    \vspace{-1.8mm}
    \subfigure[]{
    \includegraphics[scale=0.11]{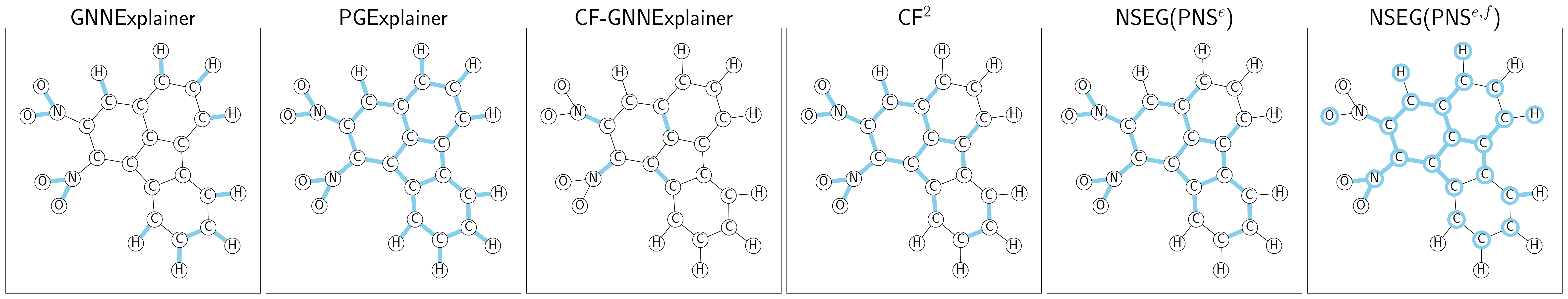}
    }
    \vspace{-1.8mm}
    \subfigure[]{
    \includegraphics[scale=0.11]{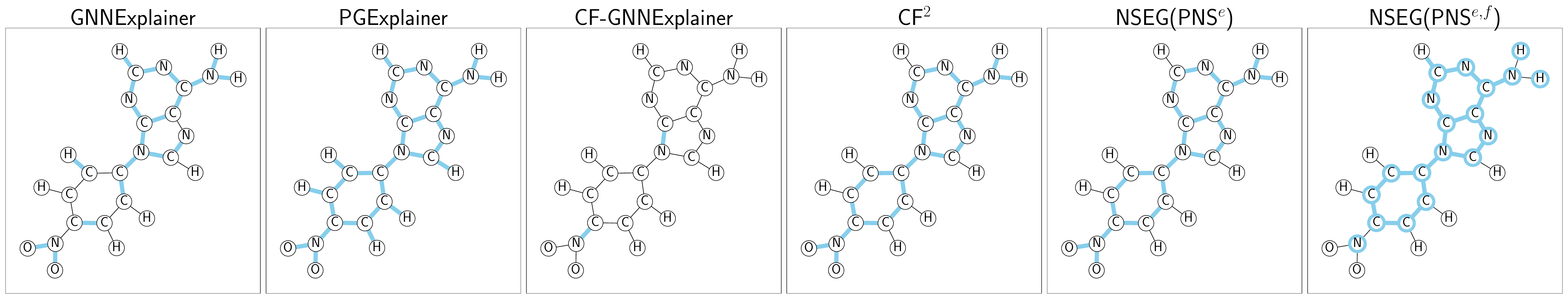}
    }
    \vspace{-1.8mm}
    \subfigure[]{
    \includegraphics[scale=0.11]{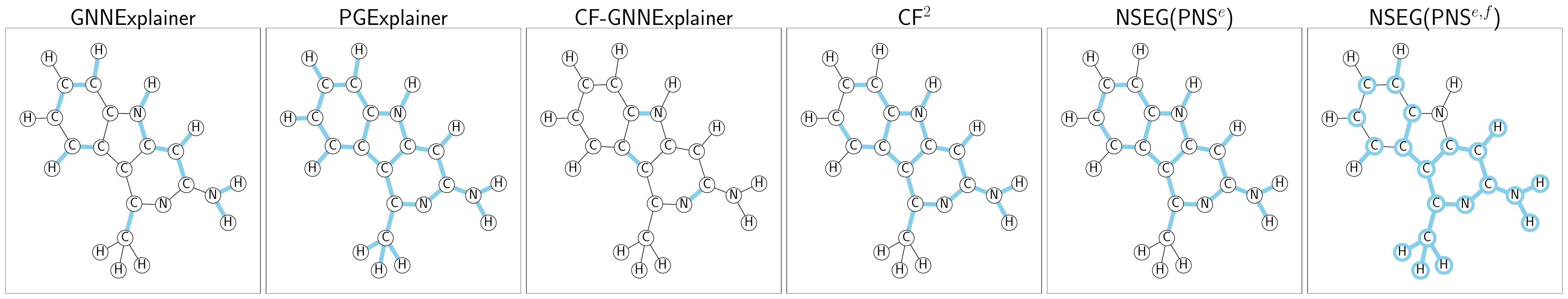}
    }
    % \vspace{-1.8mm}
    % \subfigure[]{
    % \includegraphics[scale=0.15]{Mutagenicity_82_0_0.pdf}
    % }
    \vspace{-1mm}
    \caption{Explanations of GNNExplainer, PGExplainer, CF-GNNExplainer, CF$^2$, NSEG(PNS$^{e}$), and NSEG(PNS$^{e,f}$) on various Mutagenicity instances obtained by threshold, where the explanations are highlighted in \textcolor{SkyBlue}{blue}.}
    \label{fig:more_case_mutag}
\end{figure*}

\begin{figure*}
    \centering
    % \subfigure[]{
    % \includegraphics[scale=0.45]{appendix_msrc_4-crop.pdf}
    % }
    \subfigure[]{
    \includegraphics[scale=0.32]{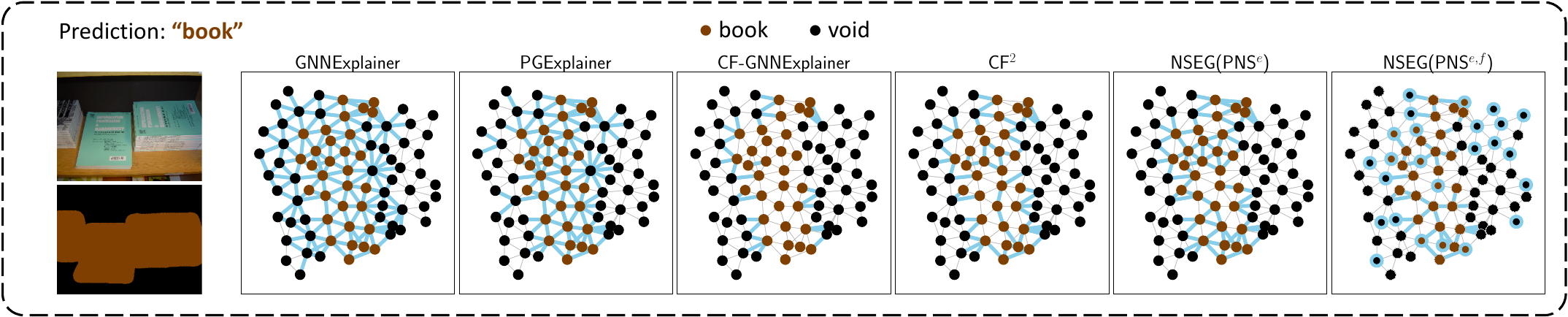}
    }
    \subfigure[]{
    \includegraphics[scale=0.32]{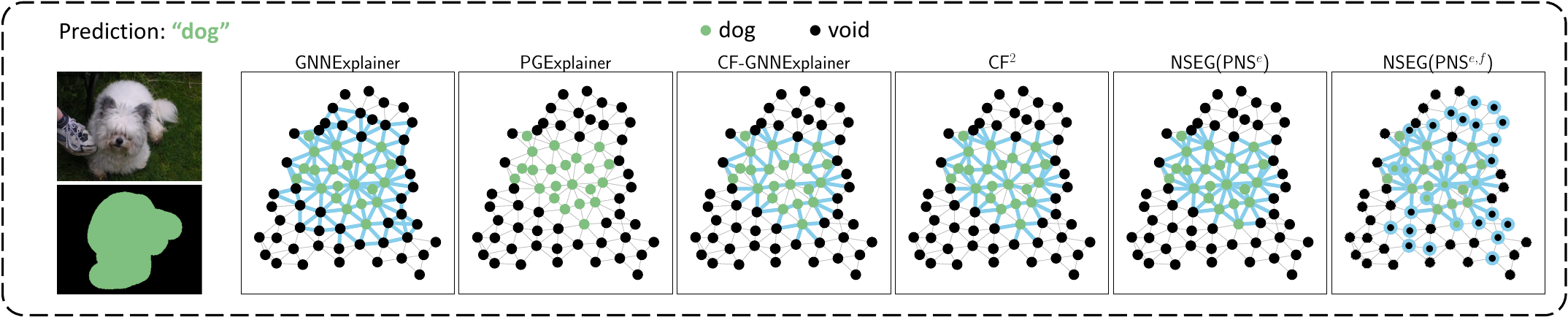}
    }
    \subfigure[]{
    \includegraphics[scale=0.32]{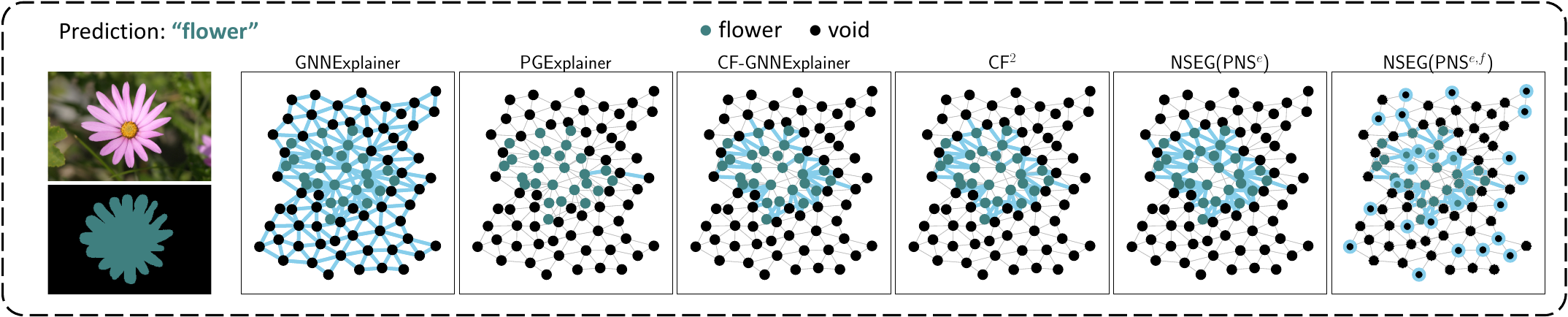}
    }
    \subfigure[]{
    \includegraphics[scale=0.32]{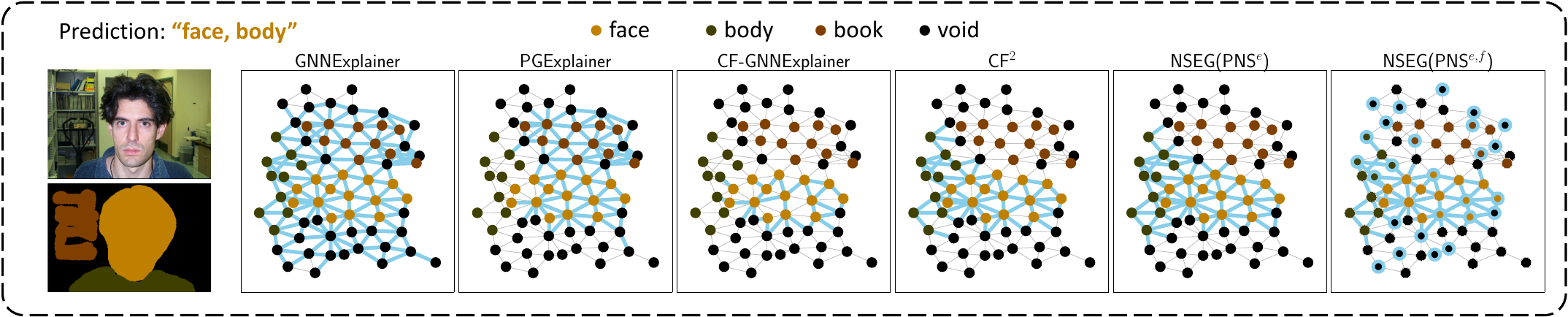}
    }

    \caption{Explanations of GNNExplainer, PGExplainer, CF-GNNExplainer, CF$^2$, NSEG(PNS$^{e}$), and NSEG(PNS$^{e,f}$) on various MSRC\_21 instances obtained by threshold, where the explanations are highlighted in \textcolor{SkyBlue}{blue}.}
    \label{fig:more_case_msrc}
\end{figure*}

We further illustrate the necessary and sufficient explanation obtained by our approach through visualization in more case studies. The visualization of explanations of Mutagenicity is shown in Figure~\ref{fig:more_case_mutag}, while the visualization of explanations of MSRC\_21 is shown in Figure~\ref{fig:more_case_msrc}.

%% References


% \bibliographystyle{elsarticle-harv}
% \bibliography{main}








% \end{thebibliography}
\end{document}